\documentclass{article}
\pdfoutput=1
\usepackage{arxiv}
\usepackage{amsmath, amssymb}
\usepackage{graphicx}
\usepackage[utf8]{inputenc}
\usepackage{hyperref}
\usepackage{multirow}
\usepackage{cleveref}
\usepackage{subfigure}
\usepackage[export]{adjustbox}
\usepackage{xspace}
\usepackage[autolanguage]{numprint}
\usepackage{todonotes}
\usepackage{chngpage}

\usepackage{xcolor}
\definecolor{GoogleGreen}{HTML}{66AA00}

\usepackage{array}
\newcolumntype{L}[1]{>{\raggedright\let\newline\\\arraybackslash\hspace{0pt}}m{#1}}
\newcolumntype{C}[1]{>{\centering\let\newline\\\arraybackslash\hspace{0pt}}m{#1}}
\newcolumntype{R}[1]{>{\raggedleft\let\newline\\\arraybackslash\hspace{0pt}}m{#1}}

\newcommand{\textfixedbf}[1]{\fontseries{b}\selectfont #1}

\newcommand{\Tone}{T1\xspace}
\newcommand{\Tonec}{T1c\xspace}
\newcommand{\Ttwo}{T2\xspace}

\begin{document}
\title{A Unified Representation Network for Segmentation with Missing Modalities}
%
\author{
  Kenneth Lau \\
  Research and Physics\\
  Elekta\\
  \texttt{kenneth.lau@elekta.com} \\
   \And
 Jonas Adler \\
  Department of Mathematics\\
  KTH - Royal institute of Technology\\
  Research and Physics\\
  Elekta\\
  \texttt{jonasadl@kth.se} \\
   \AND
  Jens Sjölund \\
  Research and Physics\\
  Elekta\\
  \texttt{jens.sjolund@elekta.com} \\
}

%
%
%
\maketitle              
\begin{abstract}
Over the last few years machine learning has demonstrated groundbreaking results in many areas of medical image analysis, including segmentation. A key assumption, however, is that the train- and test distributions match. We study a realistic scenario where this assumption is clearly violated, namely segmentation with missing input modalities.
We describe two neural network approaches that can handle a variable number of input modalities. The first is modality dropout: a simple but surprisingly effective modification of the training. The second is the unified representation network: a network architecture that maps a variable number of input modalities into a unified representation that can be used for downstream tasks such as segmentation.
We demonstrate that modality dropout makes a standard segmentation network reasonably robust to missing modalities, but that the same network works even better if trained on the unified representation.

\end{abstract}

\section{Introduction}
Image segmentation is a routine part of many diagnostic and therapeutic procedures, including radiotherapy planning. There, the required segmentations of tumors and organs at risk are often laboriously drawn by experts but still exhibit high variability, both between users and for the same user \cite{Nelms2012}. 
The prospects of improved consistency and large time-savings have attracted much attention to automatic segmentation methods \cite{Litjens2017}. But thanks to recent breakthroughs, we're finally at the cusp of widespread clinical use. The key enabler---which has virtually revolutionized medical image analysis in the last decade---is machine learning, and deep learning in particular. 

Machine learning algorithms for segmentation are typically trained as function approximations from a well-defined input to a desired output. Given enough high-quality training data, the resulting performance can be remarkable \cite{Litjens2017}. But, clinical reality is not as neat as our controlled experiments (or grand challenges for that matter \cite{menze2015multimodal}). A new frontier is to design machine learning algorithms that are resilient to violations of the underlying assumptions. This is where our contribution is at: namely segmentation with missing input modalities. 

We describe two neural network approaches that can handle a variable number of input modalities, allowing us to sidestep the combinatorial explosion that  otherwise results from training separate networks for every combination of inputs. To clarify, we don't delve into what network is best for a segmentation task with a given set of inputs---we consider a standard U-net \cite{Ronneberger2015} a good enough baseline. Instead, we first explore a simple (but surprisingly effective) modification of the training that provides robustness to missing modalities: modality dropout. Second, we describe a network architecture that uses U-nets as building blocks that map each input channel individually to a unified representation. This unified representation can then be used for downstream tasks such as segmentation.

With one notable exception \cite{Havaei2016}, most proposed approaches to segmentation with missing modalities aim to synthesize the missing modalities while leaving downstream tasks unchanged. Image-to-image translation \cite{zhang2018translating} might seem like a natural approach, but leads to the combinatorial explosion mentioned before. Better then is modality dropout, i.e. randomly removing modalities from the input during training, which has been found to improve synthesis of MR images \cite{van2019learning} but also has a longer history in other domains \cite{neverova2016moddrop,ngiam2011multimodal}. 

An alternative to modality dropout is to learn a mapping from the different input modalities to a shared representation (i.e. representation learning). The general formula for representation learning with a neural network is to have encoding and decoding steps that are both trainable. This formula is extremely versatile and applications abound: e.g. for text \cite{Radford2019}, images \cite{ngiam2011multimodal} and combinations thereof \cite{kiros2014unifying}. In fact, part of the recent successes in natural language processing have been attributed to the power of learning a unified representation through unsupervised pre-training \cite{Radford2019}. Similar approaches have also been used to synthesize medical images \cite{chartsias2017multimodal}. But in contrast to these, our philosophy is to perform unsupervised pre-training on one or several tasks, of which image synthesis is just one possibility. Then, downstream tasks such as segmentation are trained directly on the unified representation, and not on the synthesized images.


In summary, our contributions are as follows:
\begin{itemize}
    \item We demonstrate (on BRATS) that modality dropout makes a standard segmentation network (U-net) reasonably robust to missing modalities.
    \item We describe the unified representation network (URN), which learns to map a variable number of input modalities into a unified representation. We show that the standard segmentation network works even better when trained on this unified representation than with modality dropout alone.
    \item We use the unified representation network for unsupervised pre-training on two datasets with only partially overlapping modalities (BRATS and HCP), resulting in better segmentation performance.
\end{itemize}

\section{Methods}
\subsection{Baseline}
The U-net that we use as a building block throughout is a 2D, fully convolutional, U-net as originally described \cite{Ronneberger2015}, except that we use batch normalization, leaky-relu activations and only one convolutional layer at each resolution level. On segmentation tasks, we use a simple cross-entropy loss.

\subsection{Modality dropout}
By modality dropout we mean randomly zeroing out entire input channels (ordinary dropout acts on voxels). A network trained with modality dropout is forced to learn to compensate for missing modalities, presumably through features shared across modalities. This makes the performance degrade gracefully when modalities are missing at test time, as opposed to a conventionally trained network for which such a large train-test discrepancy could wreck the prediction.

We implemented modality dropout as follows. For every sample, we first determine the number $k$ of modalities to drop. We wanted the probability $p_\theta(k)$ to fall off exponentially, i.e. $p_\theta(k)\propto \theta^k$, for some constant $\theta\in (0,1)$. This would have lead to a geometric distribution if not for the fact that there is a maximum number $N_\text{max}$ of modalities that can be dropped. Modifying the normalizing constant accordingly, we find the truncated geometric distribution  
\begin{equation}
    p_\theta(k)=\frac{(1-\theta)\theta^k}{1-\theta^{N+1}},\quad k=0,\ldots,N_\text{max}.\label{eq:dropProb}
\end{equation}
Based on this equation, it is easy to sample $k$ using e.g. inverse transform sampling. Then, we select which $k$ modalities to drop uniformly at random.

\begin{figure}[t]
  \centering
  \includegraphics[width=0.75\textwidth]{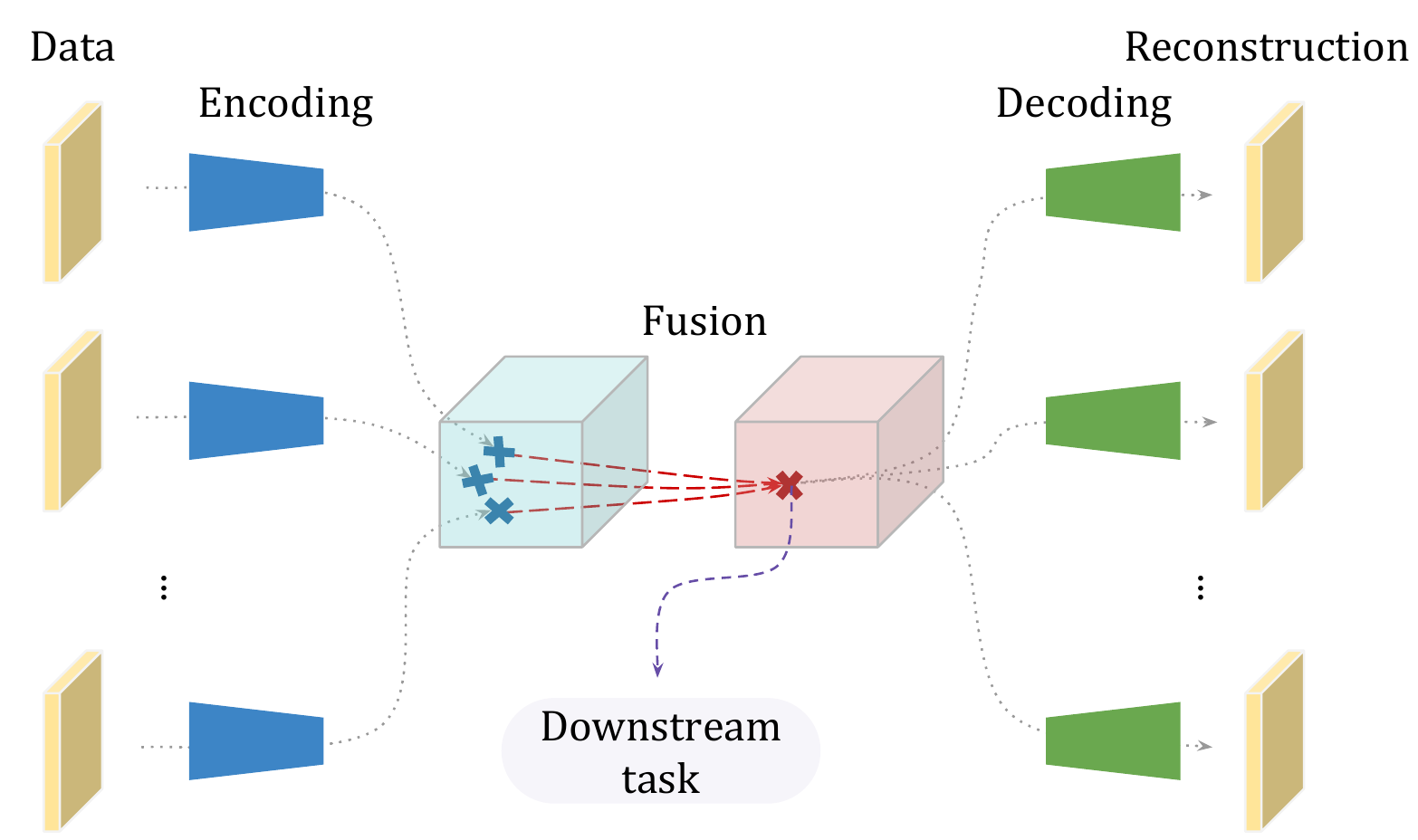}
  \caption{General architecture of a Unified Representation Network.}
  \label{fig:urn}
\end{figure}

\subsection{Unified representation network}
The unified representation network (URN) is a network architecture for learning to map a variable number of input modalities into a unified representation that downstream tasks are based on. Standard approaches, e.g. concatenation, aren't applicable in this scenario since they require a fixed set of inputs. 

A URN consists of three modules: encoding, fusion and decoding, see \cref{fig:urn}. An instance of a URN specialized for image synthesis has been proposed before \cite{chartsias2017multimodal}. The difference is that we propose to base multiple tasks directly on the unified representation---possibly even tasks it wasn't trained for. We'll now describe the three modules in more detail. 

\subsubsection{Encoding}
We use modality-specific encoders, specifically the baseline U-net described above. This implies that 
the unified representation has the same height and width as the input. We use 16 channels in the unified representation, since that's been found to work well for image synthesis \cite{chartsias2017multimodal}. An important modification is, however, to use a batch normalization with fixed parameters to standardize the channels of each encoder's output to have zero mean and unit standard deviation before fusion \cite{van2019learning}.  

\subsubsection{Fusion}
The fusion module combines the outputs $\{\tilde{z}_i\}$ of all encoders into a unified representation $z$. The main challenge is to support a variable number $n$ of inputs, which precludes concatenation and other standard approaches. When training with modality-dropout, the dropped modalities are not fused. 

It is desirable that the unified representation is an intensive property, i.e. that the magnitude of the fusion operator is independent of the number of inputs. We may ensure that $z$ is an intensive property by choosing the fusion operator to be a generalized $f$-mean,
\begin{equation}
    z = f^{-1}\left(\frac{1}{n}\sum_{i=1}^{n} f(\tilde{z}_i)\right),
\end{equation}
where $f$ is an invertible function. In this work, we stick to the mean, $f(x) = x$, for simplicity, but other familiar options include $f(x)=\exp(x)$. In principle, one could even learn $f$ with an invertible neural network \cite{Ardizzone2018}.

The fusion also has to be regularized so that all encoders map to similar representations (otherwise, the encoders are likely to learn the identity mapping). We, like others \cite{chartsias2017multimodal,van2019learning}, include the voxel-wise variance as a term in the loss function. Because of the variance computation, however, we have to limit $N$ in equation (\ref{eq:dropProb}) so that at least two modalities are always available.


\subsubsection{Decoding}
The decoding module performs a task-specific decoding based on the unified representation. Image synthesis is a task suitable for unsupervised pre-training, since it places few restrictions on the data, thereby greatly increasing the usable training data. When training to synthesize multi-modal data, we use different decoders for each modality. The decoders are shallow convolutional networks, consisting of two residual blocks and a final 1x1 convolution \cite{chartsias2017multimodal}.



\subsection{Implementation details}
The training used batch size 4, default Adam optimizer with learning rate $10^{-4}$ for segmentation and $3 \times 10^{-5}$ for pre-training. Modality dropout was done with $\theta=0.5$ for segmentation and $\theta=0.8$ for pre-training. The weighting for the similarity regularization was $10^{-4}$. All segmentation models were trained for 50 epochs. Pre-training was done until convergence. Then the segmentation was trained with these weights fixed.
The experiments were implemented in Python with TensorFlow and performed on a workstation with a NVIDIA GTX1080 Ti 11GB and an Intel Core i7-3930K CPU with 16GB RAM. Segmentation of a whole brain volume takes about 12s.

\begin{figure}[b]
  \centering
  \includegraphics[width=0.8\textwidth]{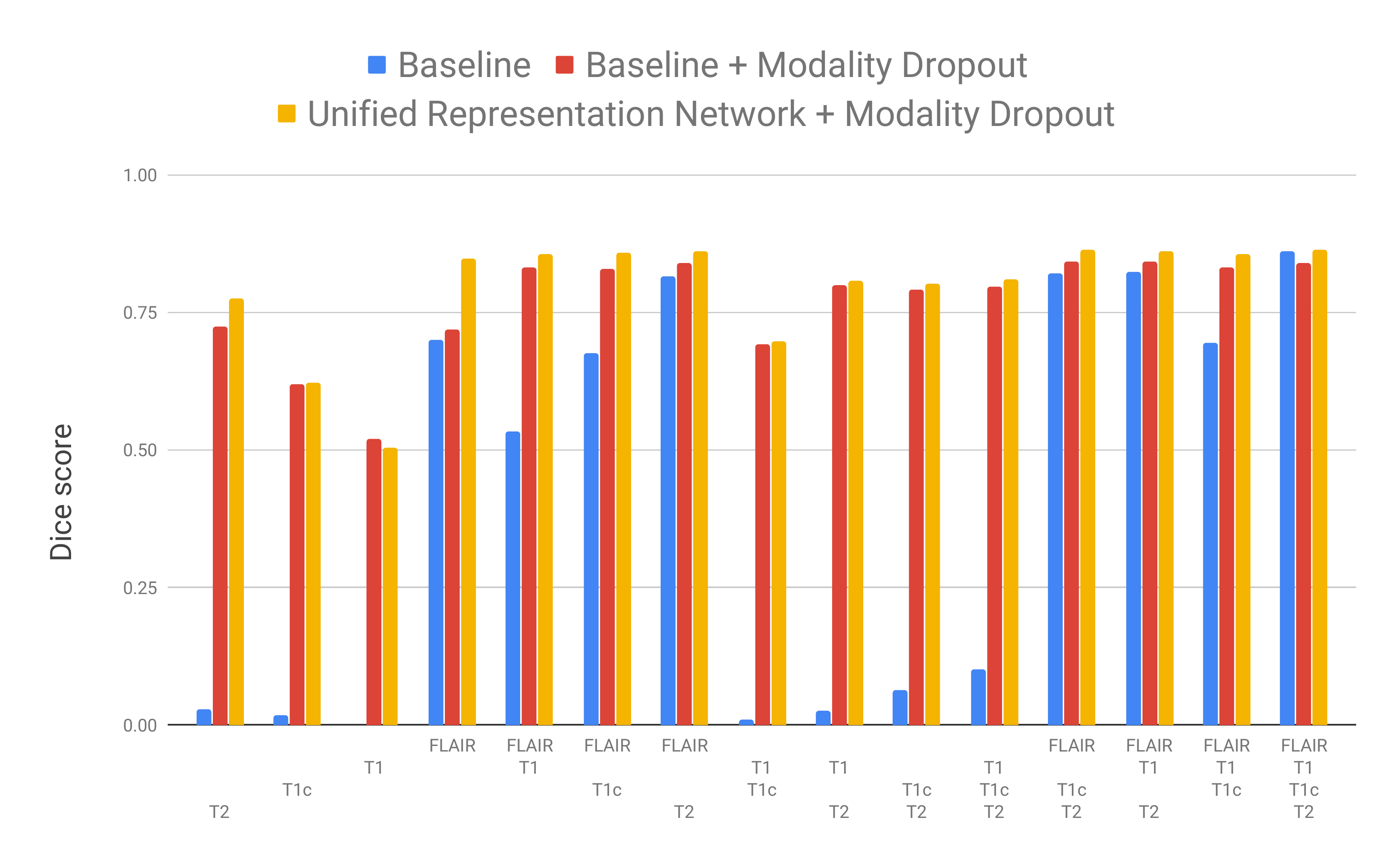}
  \caption{Dice scores of ``whole tumor'' on BRATS with different modalities.
  }
  \label{fig:Table1}
\end{figure}

\section{Experiments}
We validate the segmentation performance on BRATS. Since our main focus is to study robustness to missing modalities and not segmentation performance per se, we compare four scenarios: baseline, baseline with modality dropout, URN with modality dropout and pre-trained URN with modality dropout.

\subsection{Data}
We use MRI data from the 2018 edition of the Multimodal Brain Tumor Segmentation (BRATS) challenge \cite{Bakas2017,menze2015multimodal} which contains preprocessed MR scans (\Tone, \Ttwo, \Tonec and FLAIR) from 285 glioblastoma patients. Each voxel is labeled into one of four classes:  necrotic or non-enhancing tumor; peritumoral edema; enhancing tumor; and everything else. The evaluation, however, uses three mutually inclusive tumor regions: (i) enhancing tumor; (ii) tumor core, which includes enhancing, necrotic and non-enhancing tumor; (iii) whole tumor, which includes tumor core and peritumoral edema. The data is divided into a training set, with labels, and a validation set, without labels. We report the performance on the validation set, as evaluated by an independent online system.

For unsupervised pre-training we use MR scans (\Tone and \Ttwo) from 1108 healthy young adults provided by the Human Connectome Project (HCP) \cite{vanEssen2013} .

We preprocess all scans by bias field correcting \cite{Tustison2010} and, for every patient, independently normalizing each modality to have zero mean and unit variance within the brain mask. 
We use the same decoders in the URN as in \cite{chartsias2017multimodal} but with no activations at the final layer for image synthesis when pre-training the URN.
We split each dataset into 70\% for training and 30\% for hyperparameter tuning (validation). 
The resolutions of the MR scans are 1 mm isotropic in BRATS and 0.7 mm isotropic in HCP.

\subsection{Results}
\Cref{fig:Table1} highlights our main findings, though it only shows the Dice scores for the whole tumor and three scenarios. We excluded the URN pre-trained on pooled data from HCP and BRATS since its segmentation performance was comparable to the URN without pre-training for whole tumor which the table includes. However, the segmentation performance of enhancing tumor anc tumor core regions were substantially better. Please refer to the supplementary material for details.

As expected, the baseline U-net fails miserably most of the times a modality is missing. But, when trained with modality dropout the performance degradation is, at least for some combinations, largely offset. Even so, the URN has the highest Dice score (on whole tumor) for all combinations of inputs except one. 

Somewhat surprisingly, adding modality dropout to the baseline and the URN makes them perform better than the baseline also when all modalities are available. This could indicate that modality dropout has broader uses as a regularizer for multimodal data because it encourages feature sharing.

That some modalities are more informative than others for a given task is well-known. From \cref{fig:Table1} it appears that FLAIR is the most important for whole tumor segmentation in BRATS.   

\Cref{fig:segmentations} shows examples of segmentations generated by the different models given different combinations of inputs. The baseline (without modality dropout) completely misses the edema when FLAIR is missing, and seems quite unreliable when it comes to the necrotic and the non-enhancing tumor core as well. In the other two models, \Tonec alone is enough to generate qualitatively good segmentations of all structures despite being using mainly for defining the enhancing tumor in clinical practice.

\newcommand{\myfigure}[1]{\includegraphics[trim={1.5cm 1.2cm 1.5cm 1.5cm}, clip, width=0.18\textwidth]{#1}}

\begin{figure}[t]
\centering
\begin{tabular}{lccccc}
&
\Tonec &
\Tone, \Tonec &
\Tone, \Tonec, \Ttwo &
All &
GT \\
\raisebox{0.09\textwidth}{\rotatebox[origin=c]{90}{\centering Baseline}} &
\myfigure{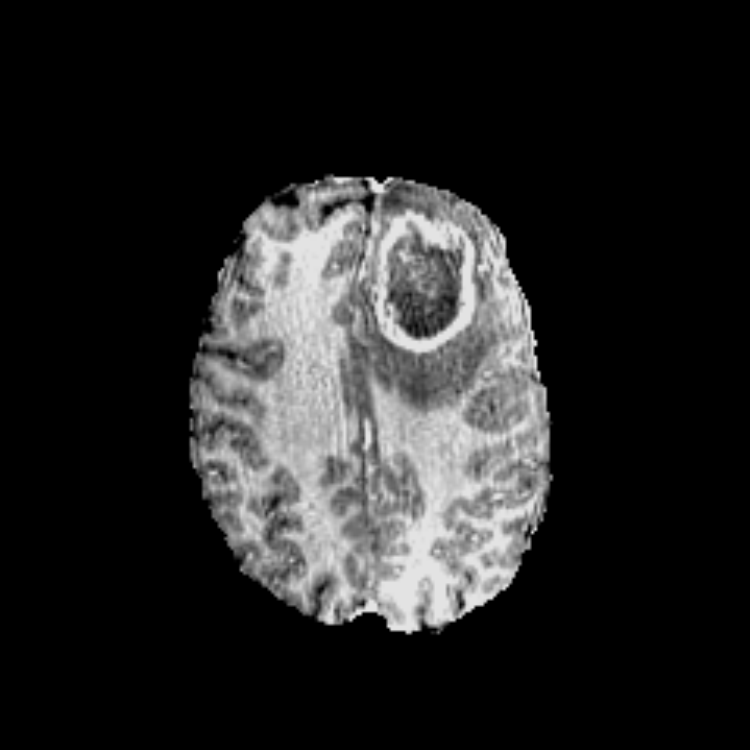} &
\myfigure{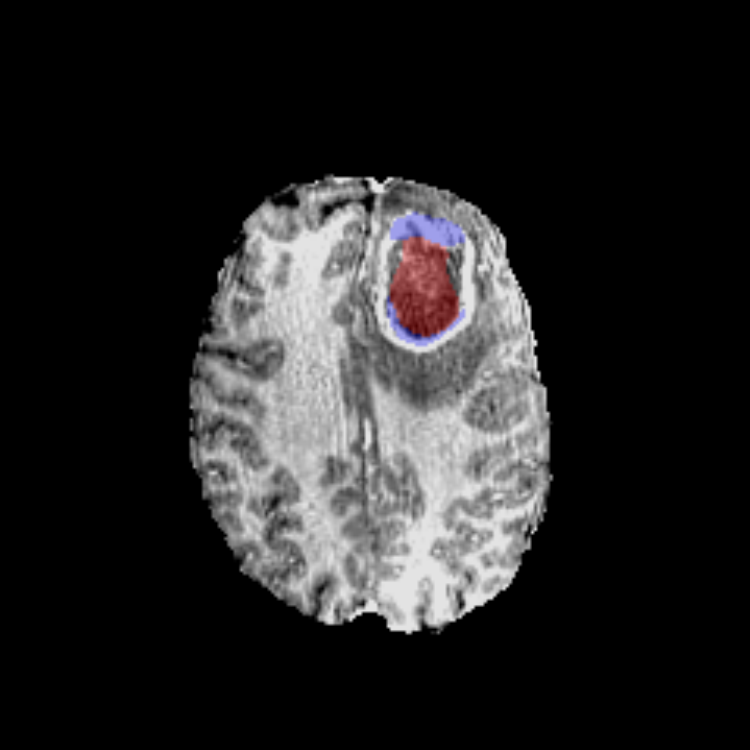} &
\myfigure{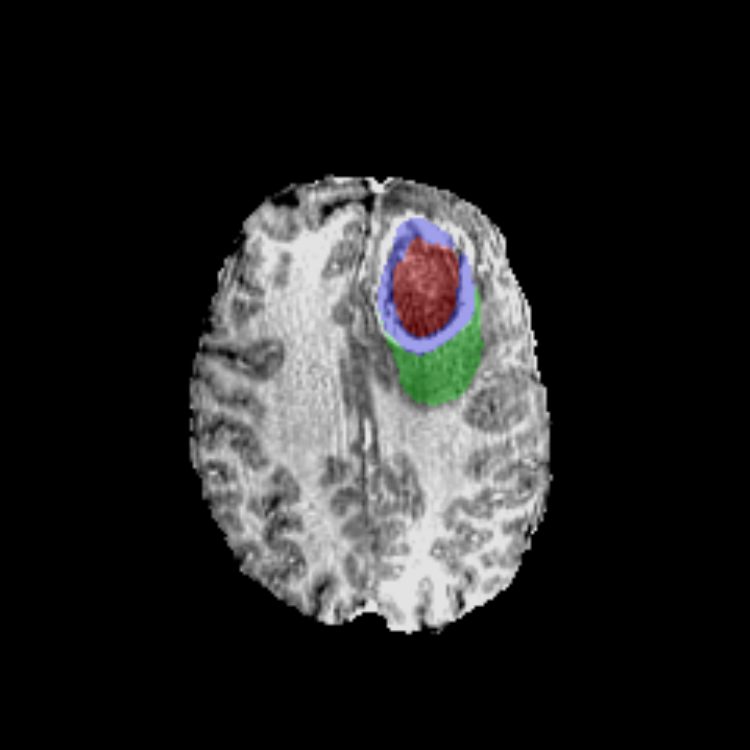} &
\myfigure{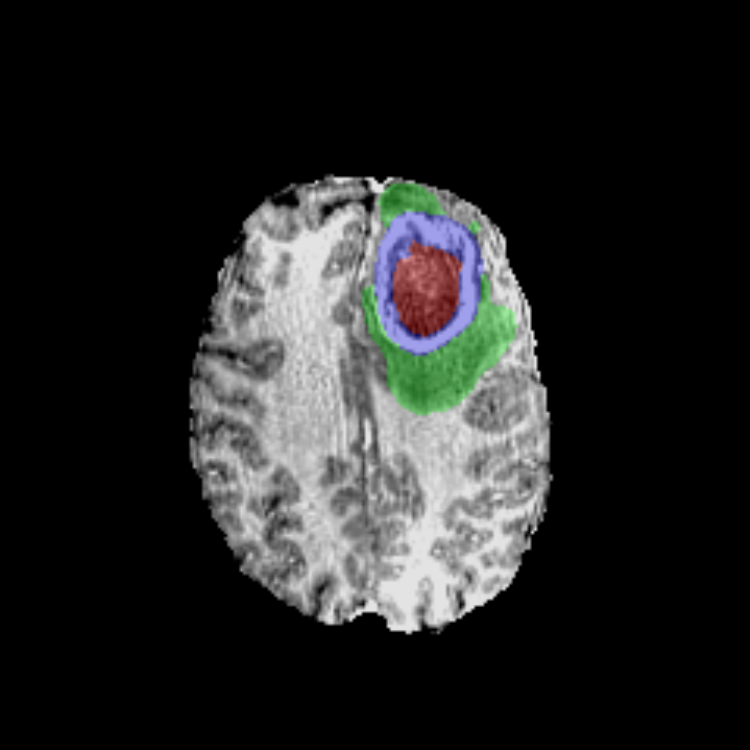} &
\myfigure{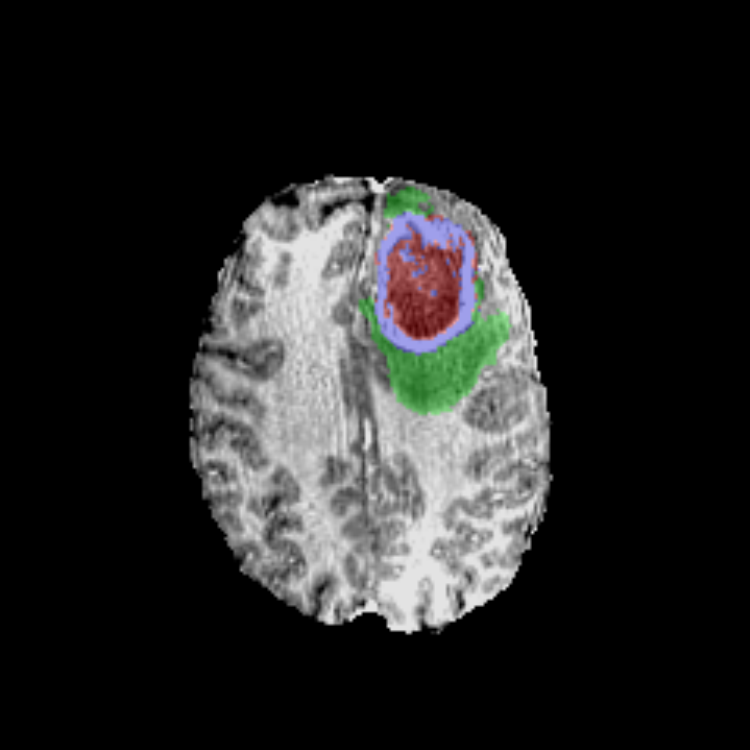}\\
\raisebox{0.09\textwidth}{\rotatebox[origin=c]{90}{Baseline + MD}} &
\myfigure{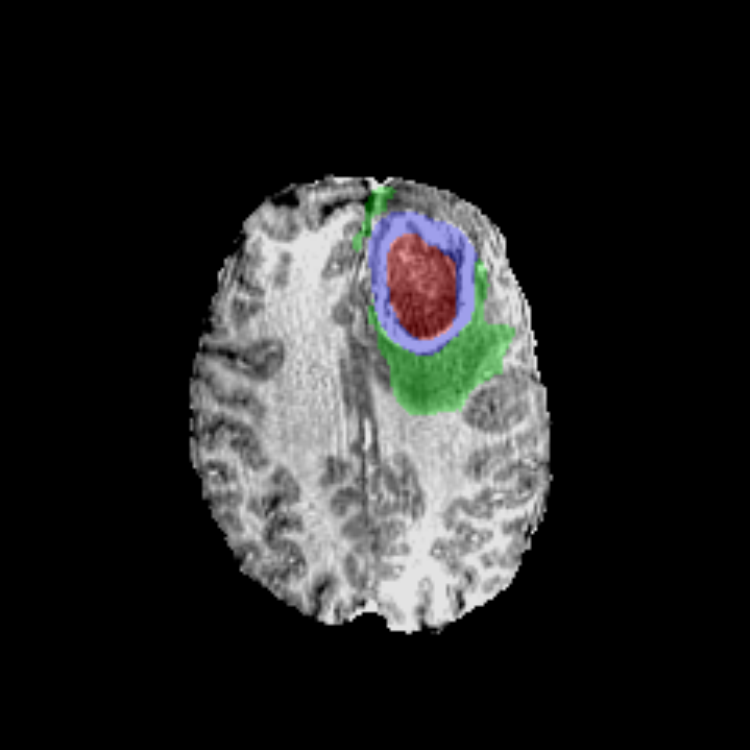} &
\myfigure{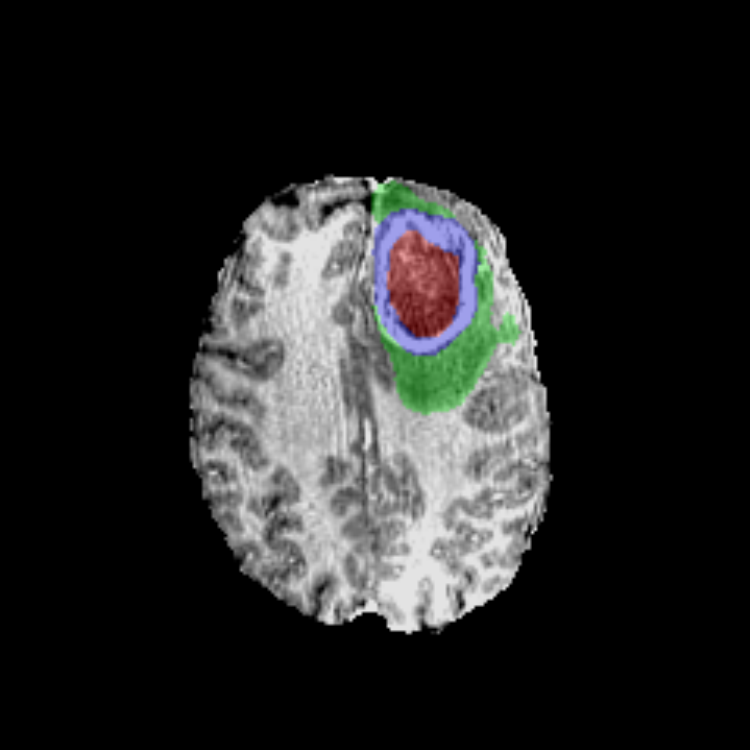} &
\myfigure{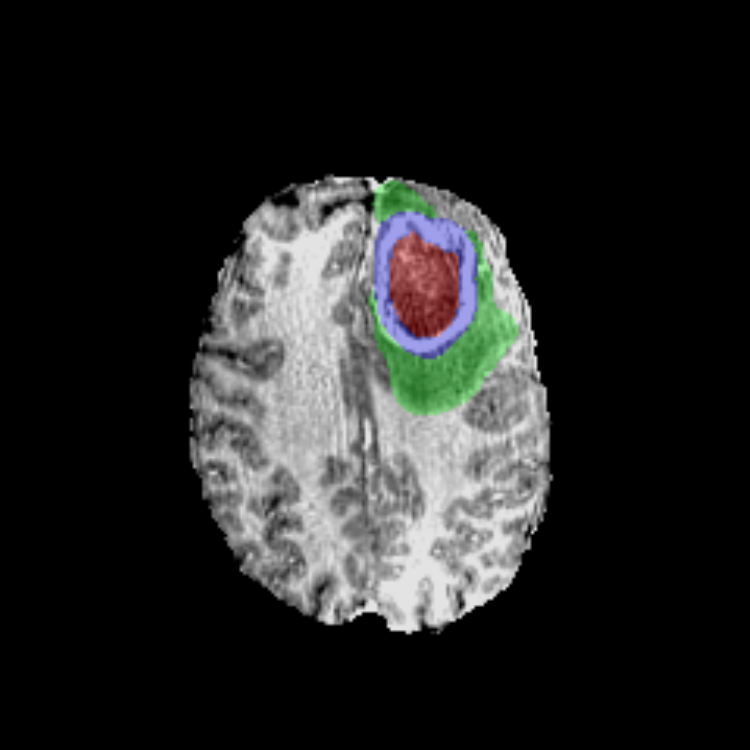} &
\myfigure{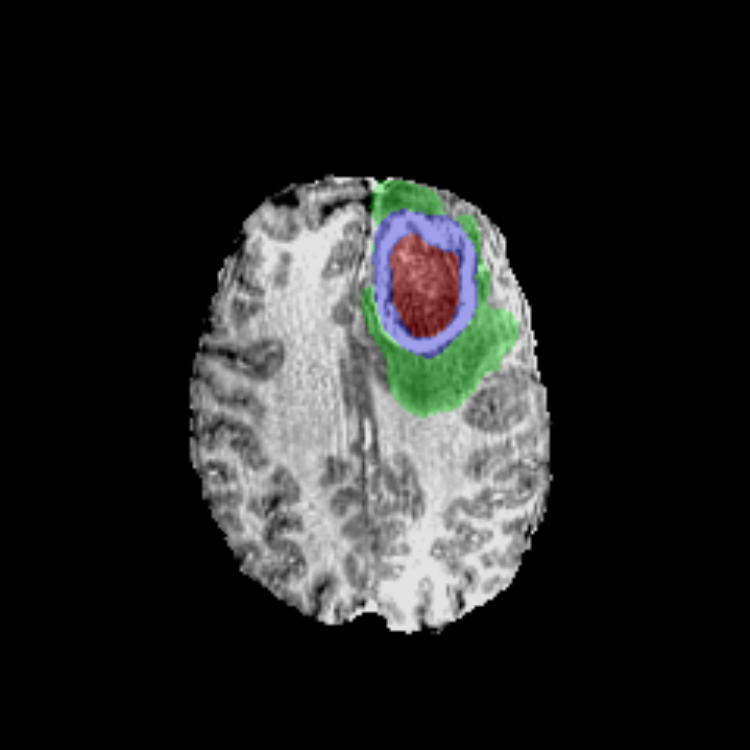} &
\myfigure{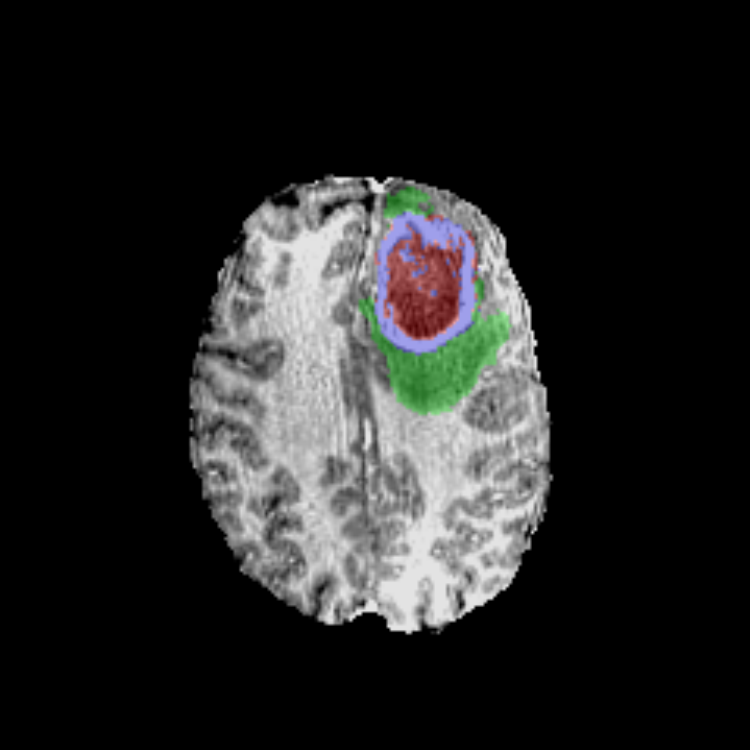}\\
\raisebox{0.09\textwidth}{\rotatebox[origin=c]{90}{URN}} &
\myfigure{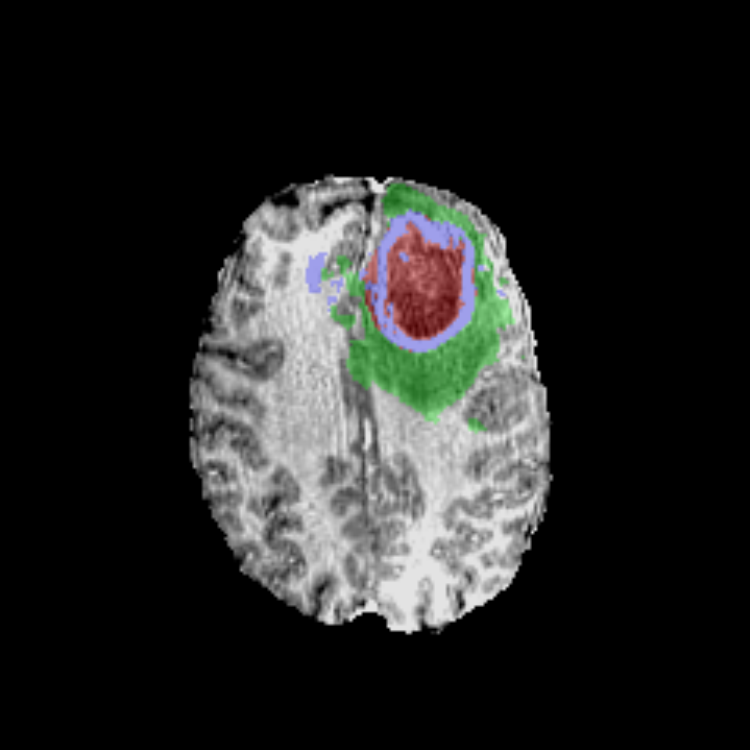} &
\myfigure{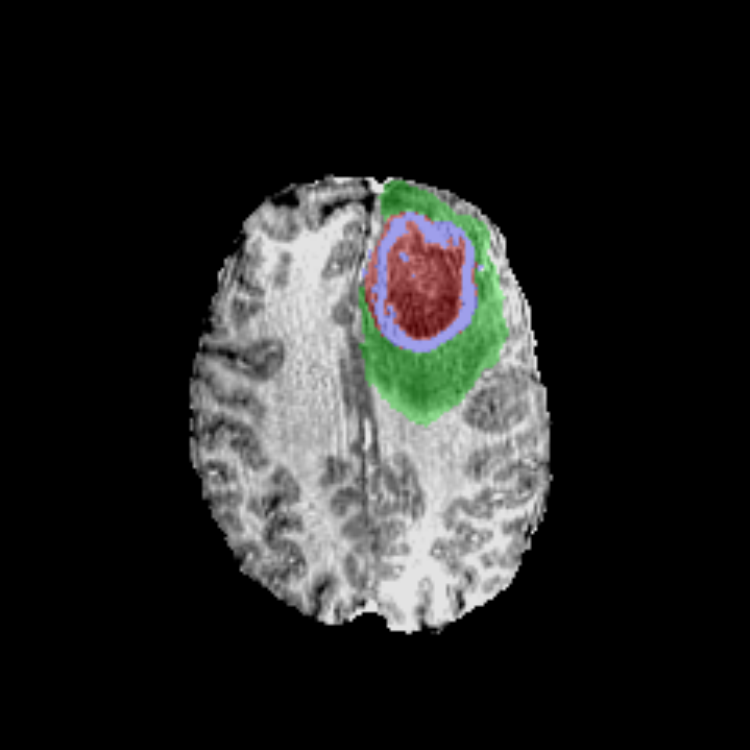} &
\myfigure{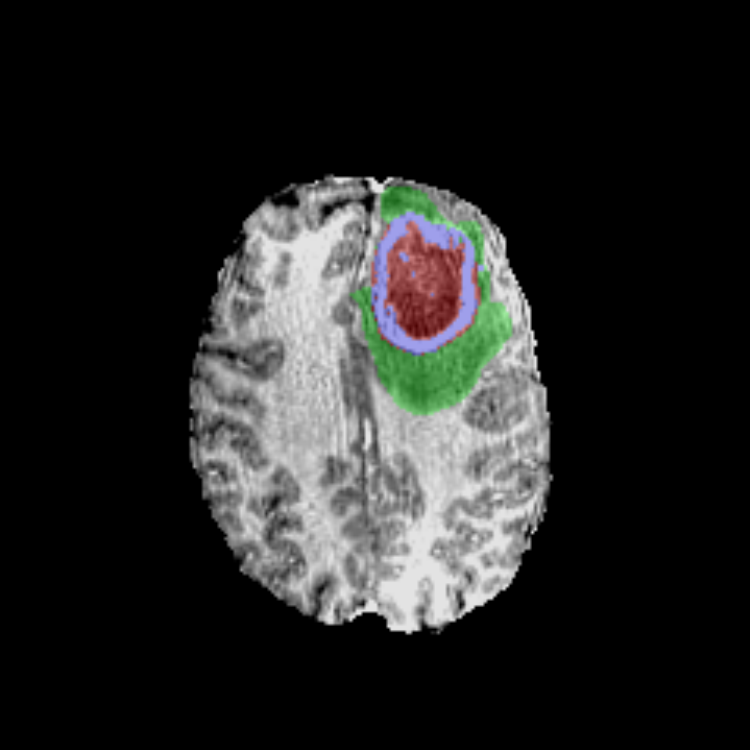} &
\myfigure{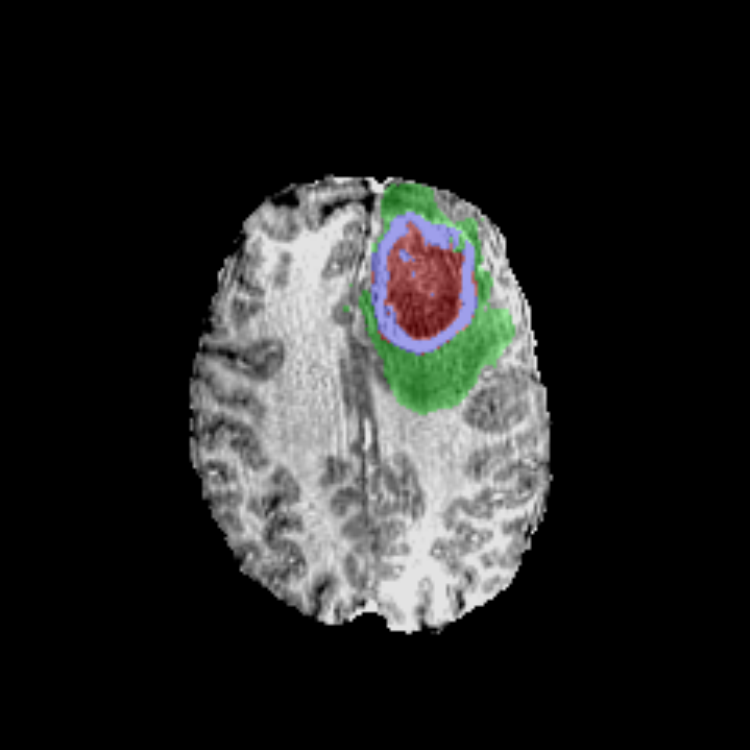} &
\myfigure{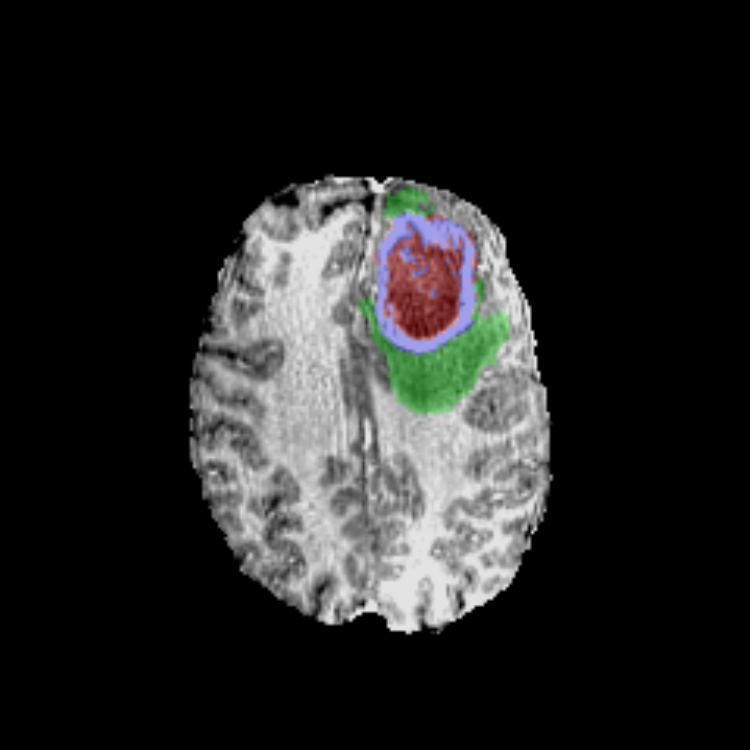}
\end{tabular}
\caption{Examples of tumor segmentations on BRATS given different combinations of inputs, overlaid on \Tonec. Red: necrotic and the non-enhancing tumor core; Green: peritumoral edema; Blue: enhancing tumor.}
\label{fig:segmentations}
\end{figure}

\section{Conclusion}
We have shown that adding modality dropout to the training of a standard segmentation network makes it robust to missing modalities at test time. We have also proposed the unified representation network (URN), which can be trained in an unsupervised manner to learn a representation useful also for other tasks. As we've shown on BRATS, the URN can be combined with modality dropout to further improve the segmentation performance when modalities are missing. Another advantage of the URN is that it allows training on multiple datasets, opening up for more powerful representation learning and possibility of improved image synthesis.

\section{Acknowledgements}

Data collection and sharing for this project was provided by the MGH-USC Human Connectome Project (HCP; Principal Investigators: Bruce Rosen, M.D., Ph.D., Arthur W. Toga, Ph.D., Van J. Weeden, MD). HCP funding was provided by the National Institute of Dental and Craniofacial Research (NIDCR), the National Institute of Mental Health (NIMH), and the National Institute of Neurological Disorders and Stroke (NINDS). HCP data are disseminated by the Laboratory of Neuro Imaging at the University of Southern California.

This research was supported by Sweden's innovation agency Vinnova, grant 2017-04596, and the Swedish Foundation for Strategic Research, grants AM13-0049 and ID14-0055.

{\small
    \bibliographystyle{splncs04}
    \bibliography{bibliography}
}

\newpage

\appendix

\section*{Supplementary material}

\subsection*{Experiments \& Results}
The following tables demonstrate the detailed results of segmentation and cross-modality image synthesis performance on BRATS with different models.

 We have also explored how unsupervised pre-training impacts the segmentation performance of a URN by comparing a URN trained to segment on only BRATS data, with a URN that pre-trained on pooled data from HCP and BRATS and segmentation is treated as a downstream task. In both cases, we use modality dropout during the training (dropped modalities are not fused).
 
 \subsubsection*{Segmentation}
 In \cref{tab:segmentation_results}, four models are compared: (1) baseline segmentation model; (2) baseline model with modality dropout (MD); (3) unified representation network with modality dropout trained only on BRATS; (4) unified representation network with modality dropout trained with BRATS and HCP. Three regions are evaluated, namely enhancing tumor (ET), whole tumor (WT) and tumor core (TC). Different combinations of inputs are denoted as available ($\bullet$) and unavailable ($\circ$). Modalities are denoted as FLAIR, \Tone, \Tonec and \Ttwo. Note: Results of \Tone using Baseline model is unable to be retrieved from the leaderboard portal.

\newcommand{\tabledigit}[1]{\nprounddigits{1}\numprint{#1}}
\newcommand{\tabledigitbold}[1]{\textfixedbf{\nprounddigits{1}\numprint{#1}}}

\begin{table}[h]
\begin{adjustwidth}{-.5in}{-.5in}  
\centering
\caption{Dice scores for different models on BRATS.}\begin{tabular}{C{0.6cm}C{0.6cm}C{0.6cm}C{0.6cm}|R{0.8cm}R{0.8cm}R{0.8cm}|R{0.8cm}R{0.8cm}R{0.8cm}|R{0.8cm}R{0.8cm}R{0.8cm}|R{0.8cm}R{0.8cm}R{0.8cm}}
\multicolumn{4}{c}{Modalities} & \multicolumn{3}{c}{Baseline} & \multicolumn{3}{c}{Baseline + MD} & \multicolumn{3}{c}{URN + MD} & \multicolumn{3}{c}{URN + MD w/ HCP} \\
\cline{1-16}
$F$ & \Tone & \Tonec & \Ttwo & ET & WT & TC & ET & WT & TC & ET & WT & TC & ET & WT & TC \\
\hline
$\bullet$ & $\bullet$ & $\bullet$ & $\bullet$ & \tabledigitbold{74.18} & \tabledigit{86.15} & \tabledigit{75.85} & \tabledigit{63.85} & \tabledigit{84.09} & \tabledigit{73.85} & \tabledigit{69.87} & \tabledigitbold{86.34} & \tabledigit{71.78} & \tabledigitbold{71.27} & \tabledigit{86.07} & \tabledigitbold{77.99} \\
$\bullet$ & $\bullet$ & $\bullet$ & $\circ$ & \tabledigit{54.84} & \tabledigit{69.34} & \tabledigit{52.51} & \tabledigit{63.18} & \tabledigit{83.20} & \tabledigit{72.95} & \tabledigit{71.01} & \tabledigitbold{85.62} & \tabledigit{72.02} & \tabledigitbold{72.29} & \tabledigit{85.25} & \tabledigitbold{77.60} \\
$\bullet$ & $\bullet$ & $\circ$ & $\bullet$ & \tabledigit{3.83} & \tabledigit{82.32} & \tabledigit{34.37} & \tabledigit{30.52} & \tabledigit{84.19} & \tabledigit{58.88} & \tabledigit{25.84} & \tabledigitbold{86.06} & \tabledigit{52.46} & \tabledigitbold{39.66} & \tabledigit{85.52} & \tabledigitbold{62.98} \\
$\bullet$ & $\circ$ & $\bullet$ & $\bullet$ & \tabledigit{66.29} & \tabledigit{82.03} & \tabledigit{67.98} & \tabledigit{63.75} & \tabledigit{84.11} & \tabledigit{74.31} & \tabledigit{69.80} & \tabledigitbold{86.51} & \tabledigit{72.16} & \tabledigitbold{71.05} & \tabledigit{85.95} & \tabledigitbold{78.84} \\
$\circ$ & $\bullet$ & $\bullet$ & $\bullet$ & \tabledigit{17.87} & \tabledigit{10.13} & \tabledigit{18.28} & \tabledigit{65.72} & \tabledigit{79.65} & \tabledigit{71.04} & \tabledigit{68.49} & \tabledigit{81.11} & \tabledigit{69.53} & \tabledigitbold{71.82} & \tabledigitbold{82.10} & \tabledigitbold{76.30} \\
$\circ$ & $\circ$ & $\bullet$ & $\bullet$ & \tabledigit{12.14} & \tabledigit{6.30} & \tabledigit{11.51} & \tabledigit{62.39} & \tabledigit{79.19} & \tabledigit{71.03} & \tabledigit{67.59} & \tabledigit{80.27} & \tabledigit{68.92} & \tabledigitbold{73.44} & \tabledigitbold{81.38} & \tabledigitbold{77.61} \\
$\circ$ & $\bullet$ & $\circ$ & $\bullet$ & \tabledigit{0.06} & \tabledigit{2.64} & \tabledigit{3.97} & \tabledigit{30.77} & \tabledigit{79.87} & \tabledigit{53.04} & \tabledigit{25.24} & \tabledigitbold{80.78} & \tabledigit{48.55} & \tabledigitbold{40.46} & \tabledigit{80.66} & \tabledigitbold{59.91} \\
$\circ$ & $\bullet$ & $\bullet$ & $\circ$ & \tabledigit{2.30} & \tabledigit{1.01} & \tabledigit{2.82} & \tabledigit{60.26} & \tabledigit{69.26} & \tabledigit{67.00} & \tabledigit{66.45} & \tabledigit{69.80} & \tabledigit{65.92} & \tabledigitbold{68.80} & \tabledigitbold{70.65} & \tabledigitbold{71.64} \\
$\bullet$ & $\circ$ & $\circ$ & $\bullet$ & \tabledigit{4.55} & \tabledigit{81.65} & \tabledigit{14.79} & \tabledigit{31.56} & \tabledigit{84.06} & \tabledigit{55.67} & \tabledigit{25.23} & \tabledigitbold{86.25} & \tabledigit{50.71} & \tabledigitbold{41.73} & \tabledigit{85.41} & \tabledigitbold{63.05} \\
$\bullet$ & $\circ$ & $\bullet$ & $\circ$ & \tabledigit{42.94} & \tabledigit{67.56} & \tabledigit{43.17} & \tabledigit{62.48} & \tabledigit{82.78} & \tabledigit{73.19} & \tabledigit{70.41} & \tabledigitbold{85.80} & \tabledigit{72.52} & \tabledigitbold{71.01} & \tabledigit{84.98} & \tabledigitbold{76.84} \\
$\bullet$ & $\bullet$ & $\circ$ & $\circ$ & \tabledigit{6.66} & \tabledigit{53.41} & \tabledigit{11.42} & \tabledigit{24.66} & \tabledigit{83.23} & \tabledigit{54.48} & \tabledigit{25.34} & \tabledigitbold{85.46} & \tabledigit{52.56} & \tabledigitbold{36.88} & \tabledigit{84.58} & \tabledigitbold{59.55} \\
$\bullet$ & $\circ$ & $\circ$ & $\circ$ & \tabledigit{6.06} & \tabledigit{70.05} & \tabledigit{0.00} & \tabledigit{16.06} & \tabledigit{71.92} & \tabledigit{28.27} & \tabledigit{23.61} & \tabledigitbold{84.76} & \tabledigit{50.40} & \tabledigitbold{31.79} & \tabledigit{84.14} & \tabledigitbold{52.11} \\
$\circ$ & $\bullet$ & $\circ$ & $\circ$ & - & - & - & \tabledigit{10.25} & \tabledigit{51.94} & \tabledigit{30.29} & \tabledigit{19.09} & \tabledigit{50.42} & \tabledigit{34.21} & \tabledigitbold{20.96} & \tabledigitbold{54.73} & \tabledigitbold{37.18} \\
$\circ$ & $\circ$ & $\bullet$ & $\circ$ & \tabledigit{5.81} & \tabledigit{1.78} & \tabledigit{3.90} & \tabledigit{48.97} & \tabledigit{61.82} & \tabledigit{58.62} & \tabledigit{55.83} & \tabledigit{62.16} & \tabledigit{58.45} & \tabledigitbold{61.09} & \tabledigitbold{63.63} & \tabledigitbold{65.80} \\
$\circ$ & $\circ$ & $\circ$ & $\bullet$ & \tabledigit{1.52} & \tabledigit{2.92} & \tabledigit{3.91} & \tabledigit{25.63} & \tabledigit{72.37} & \tabledigit{47.35} & \tabledigit{20.26} & \tabledigitbold{77.54} & \tabledigit{43.64} & \tabledigitbold{38.51} & \tabledigit{75.07} & \tabledigitbold{55.58} \\
\end{tabular}
\label{tab:segmentation_results}
\end{adjustwidth}
\end{table}

\newpage

\subsubsection*{Image synthesis}
In \cref{tab:image_synthesis}, two models are compared: (1) unified representation network with modality dropout trained with BRATS; (2) unified representation network with modality dropout trained with BRATS and HCP. The peak signal-to-noise ratio of different modalities in BRATS are evaluated. Different combinations of inputs are denoted as available ($\bullet$) and unavailable ($\circ$). Modalities are denoted as FLAIR, \Tone, \Tonec and \Ttwo.
 
\Cref{fig:URN_synthesis} and \cref{fig:URN_wHCP_synthesis} show examples of image synthesis given different inputs that are trained on different datasets. Reconstructed images with given inputs are indicated with green border for better visualization.

\begin{table}[h]
\centering
\caption{Peak signal-to-noise ratio of synthesized BRATS images.}\begin{tabular}{C{0.6cm}C{0.6cm}C{0.6cm}C{0.6cm}|C{0.7cm}C{0.7cm}C{0.7cm}C{0.7cm}|C{0.7cm}C{0.7cm}C{0.7cm}C{0.7cm}}
\multicolumn{4}{c}{Modalities} & \multicolumn{4}{c}{URN} & \multicolumn{4}{c}{URN w/ HCP} \\
\cline{1-12}
$F$ & \Tone & \Tonec & \Ttwo & $F$ & \Tone & \Tonec & \Ttwo & $F$ & \Tone & \Tonec & \Ttwo \\
\hline
$\bullet$ & $\bullet$ & $\bullet$ & $\circ$ & - & - & - & \textfixedbf{19.8} & - & - & - & 18.6 \\
$\bullet$ & $\bullet$ & $\circ$ & $\bullet$ & - & - & \textfixedbf{19.3} & - & - & - & 19.1 & - \\
$\bullet$ & $\circ$ & $\bullet$ & $\bullet$ & - & 22.3 & - & - & - & \textfixedbf{22.4} & - & - \\
$\circ$ & $\bullet$ & $\bullet$ & $\bullet$ & \textfixedbf{18.7} & - & - & - & 18.4 & - & - & - \\
$\circ$ & $\circ$ & $\bullet$ & $\bullet$ & \textfixedbf{18.1} & \textfixedbf{21.7} & - & - & 18.0 & 21.5 & - & - \\
$\circ$ & $\bullet$ & $\circ$ & $\bullet$ & \textfixedbf{18.7} & - & \textfixedbf{19.2} & - & 18.4 & - & 19.0 & - \\
$\circ$ & $\bullet$ & $\bullet$ & $\circ$ & \textfixedbf{17.2} & - & - & \textfixedbf{18.2} & 17.0 & - & - & 17.3 \\
$\bullet$ & $\circ$ & $\circ$ & $\bullet$ & - & 20.1 & \textfixedbf{17.5} & - & - & \textfixedbf{20.2} & 17.2 & - \\
$\bullet$ & $\circ$ & $\bullet$ & $\circ$ & - & \textfixedbf{21.7} & - & \textfixedbf{19.1} & - & 21.5 & - & 18.4 \\
$\bullet$ & $\bullet$ & $\circ$ & $\circ$ & - & - & \textfixedbf{19.0} & \textfixedbf{19.4} & - & - & 18.8 & 18.4 \\
$\bullet$ & $\circ$ & $\circ$ & $\circ$ & - & 5.7 & \textfixedbf{8.0} & 6.2 & - & \textfixedbf{15.2} & 7.2 & \textfixedbf{14.1} \\
$\circ$ & $\bullet$ & $\circ$ & $\circ$ & \textfixedbf{16.5} & - & \textfixedbf{16.0} & \textfixedbf{17.5} & 13.3 & - & 15.7 & 16.4 \\
$\circ$ & $\circ$ & $\bullet$ & $\circ$ & 14.1 & 16.1 & - & \textfixedbf{16.3} & \textfixedbf{16.1} & \textfixedbf{17.1} & - & 16.0 \\
$\circ$ & $\circ$ & $\circ$ & $\bullet$ & 15.1 & 18.0 & 14.3 & - & \textfixedbf{16.5} & \textfixedbf{18.6} & \textfixedbf{15.9} & - \\
\end{tabular}
\label{tab:image_synthesis}
\end{table}

\newcommand{\synfigure}[1]{\includegraphics[trim={1.5cm 1.2cm 1.5cm 1.5cm}, clip, width=0.17\textwidth, cfbox=white 1.5pt 0pt]{#1}}
\newcommand{\synfiguresyntetic}[1]{\includegraphics[trim={1.5cm 1.2cm 1.5cm 1.5cm}, clip, width=0.17\textwidth, cfbox=GoogleGreen 1.5pt 0pt]{#1}}

\begin{figure}[t]
\centering
\begin{tabular}{lccccc}
&
\Tonec &
\Tone, \Tonec &
\Tone, \Tonec, \Ttwo &
All &
GT \\
\raisebox{0.09\textwidth}{\rotatebox[origin=c]{90}{\centering FLAIR}} &
\synfigure{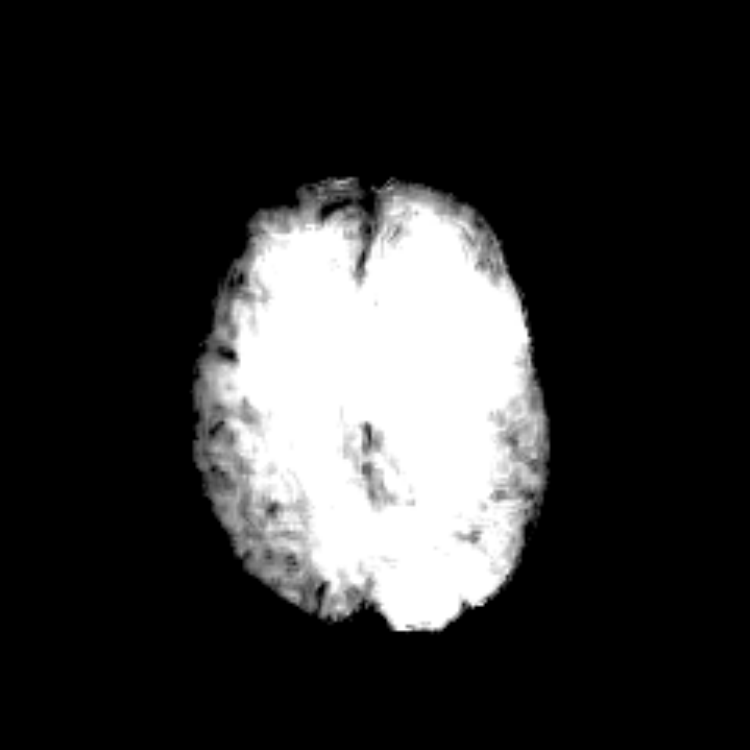} &
\synfigure{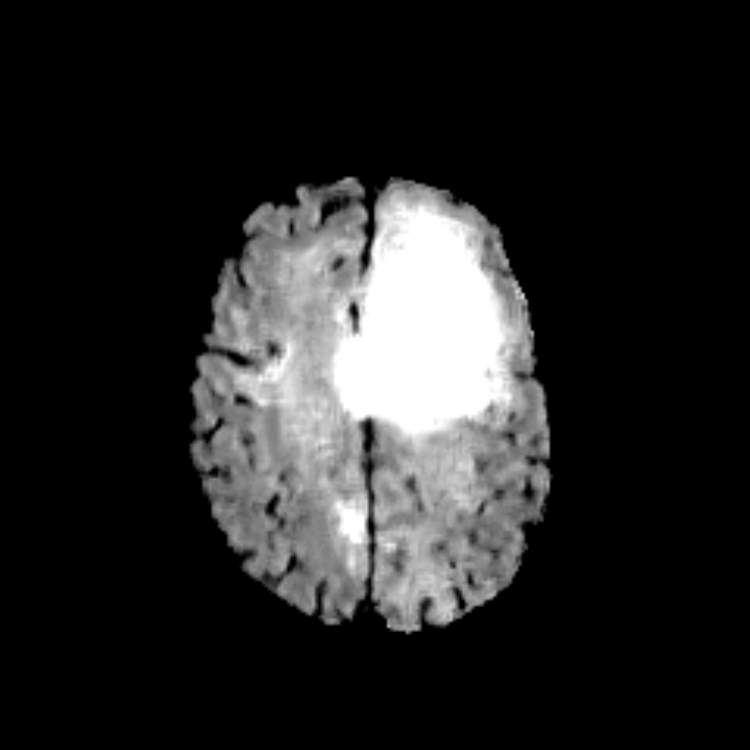} &
\synfigure{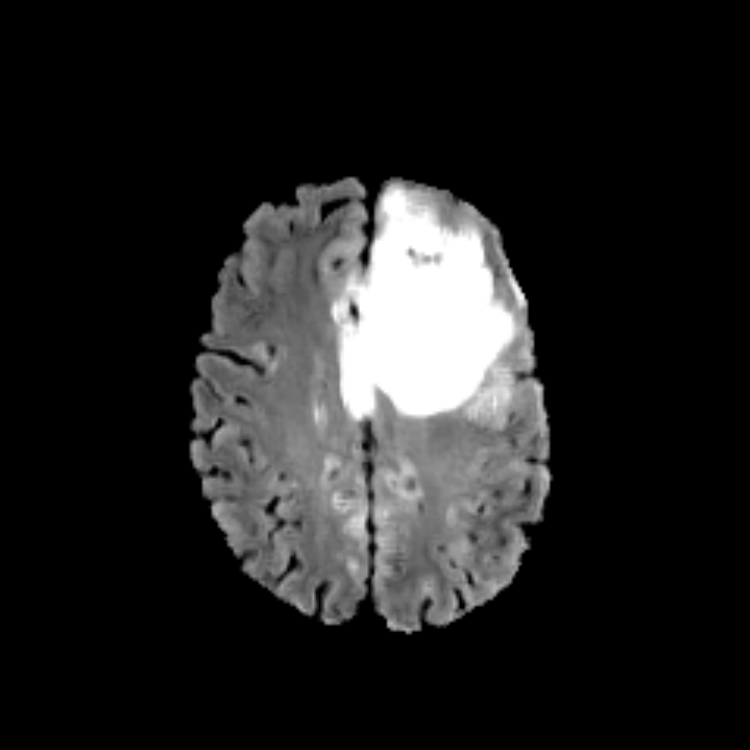} &
\synfiguresyntetic{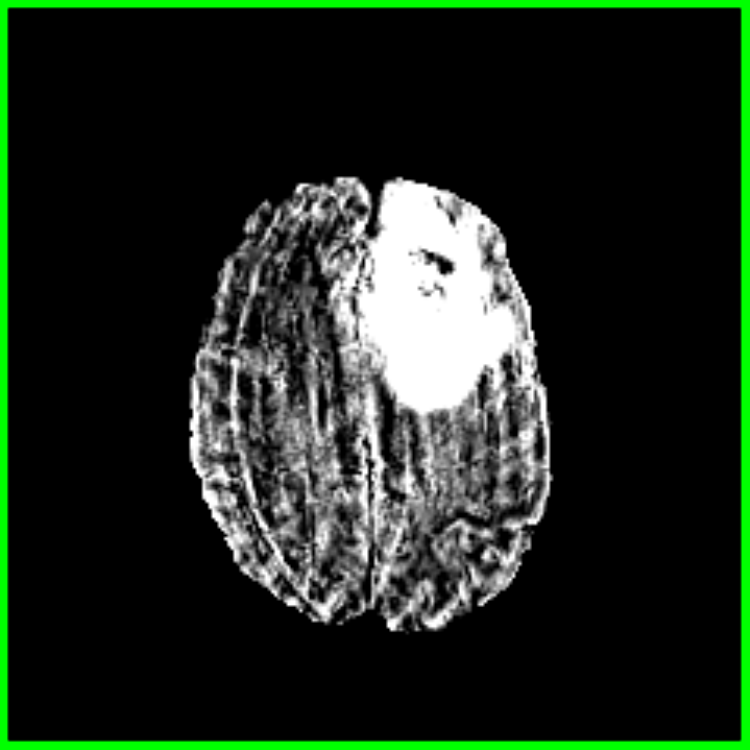} &
\synfigure{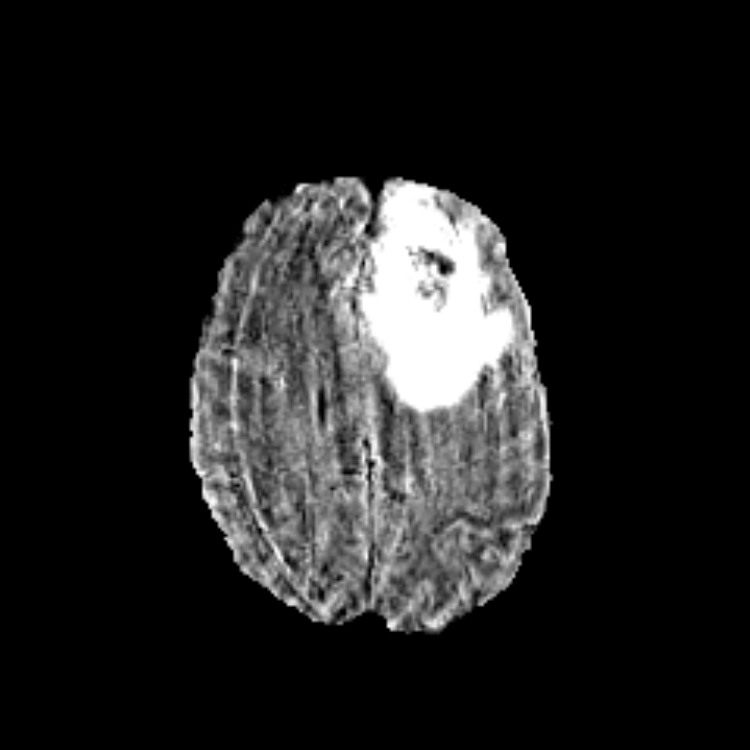}\\
\raisebox{0.09\textwidth}{\rotatebox[origin=c]{90}{\Ttwo}} &
\synfigure{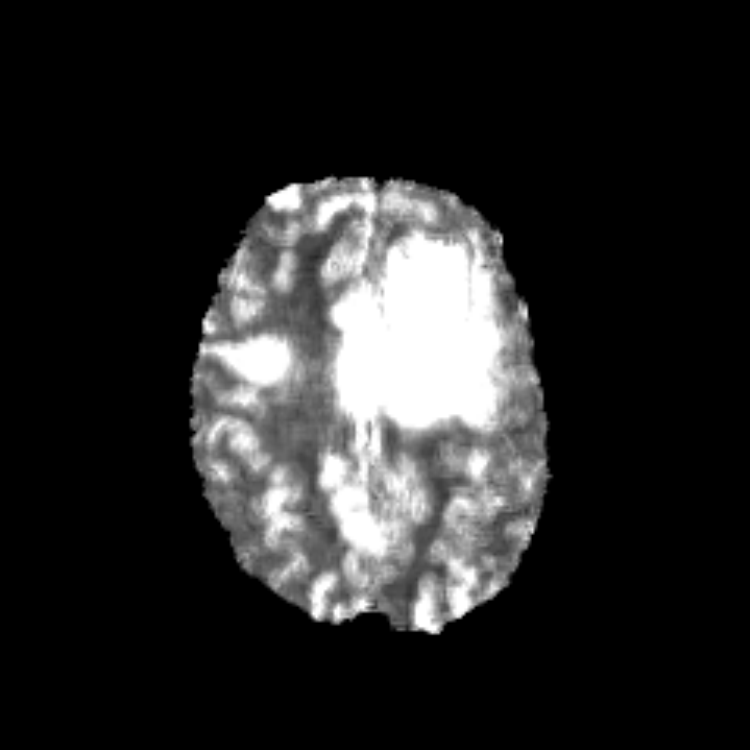} &
\synfigure{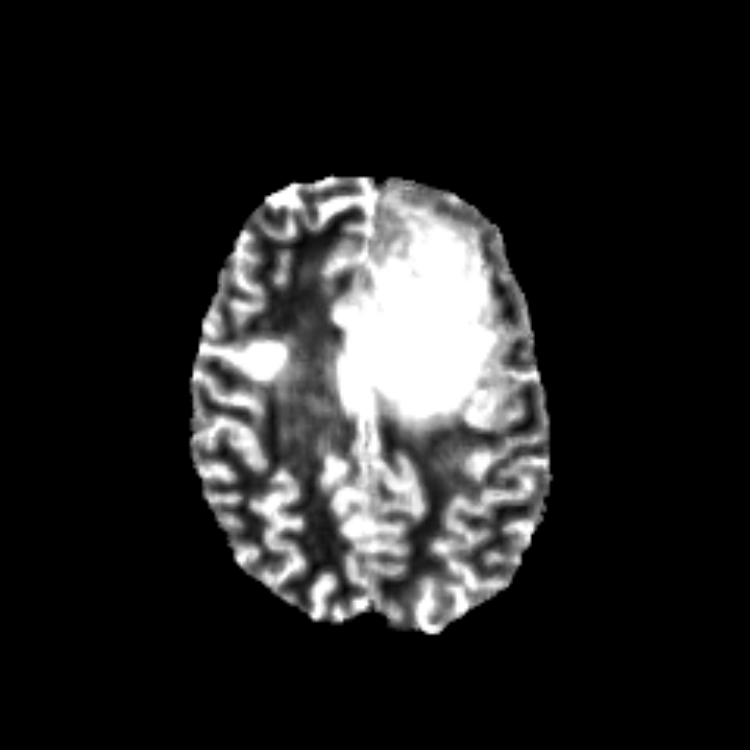} &
\synfiguresyntetic{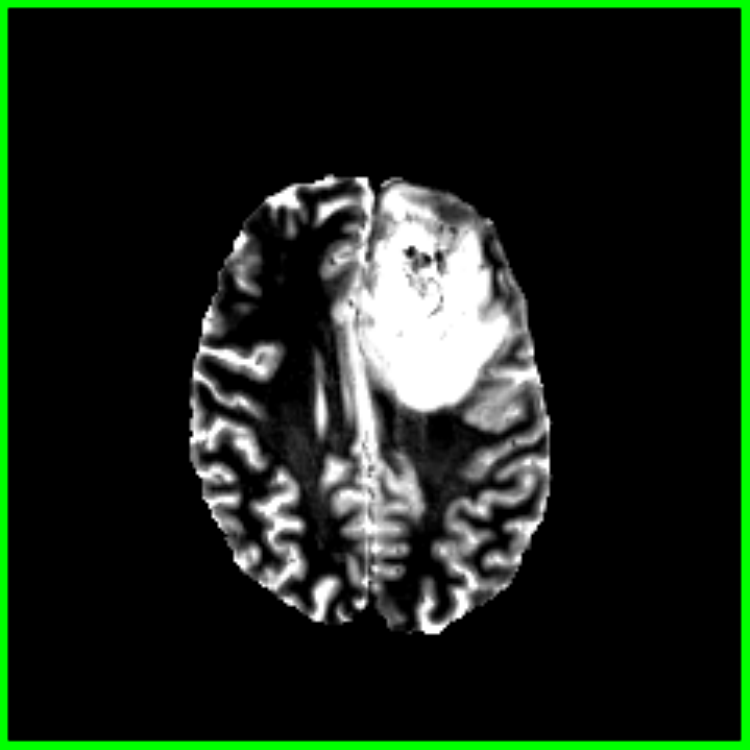} &
\synfiguresyntetic{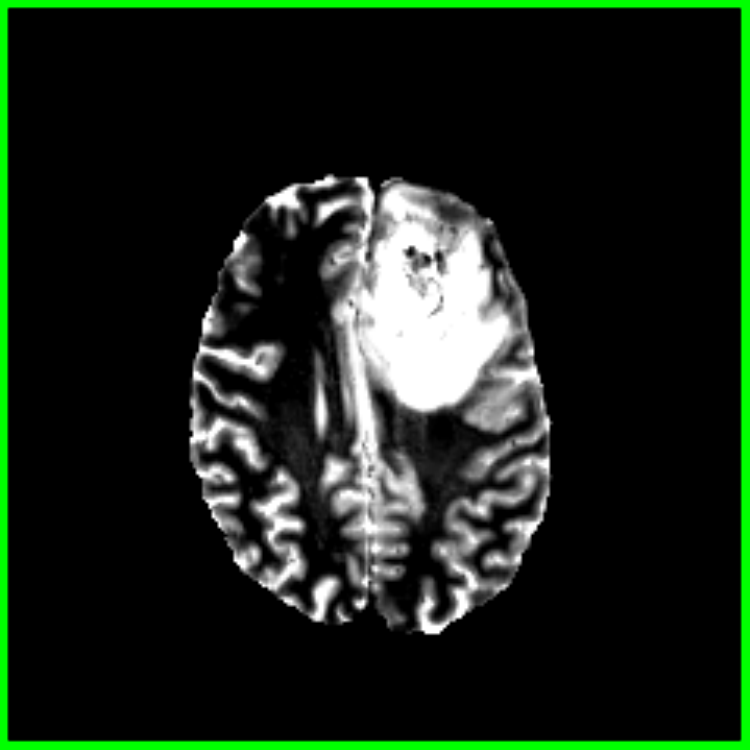} &
\synfigure{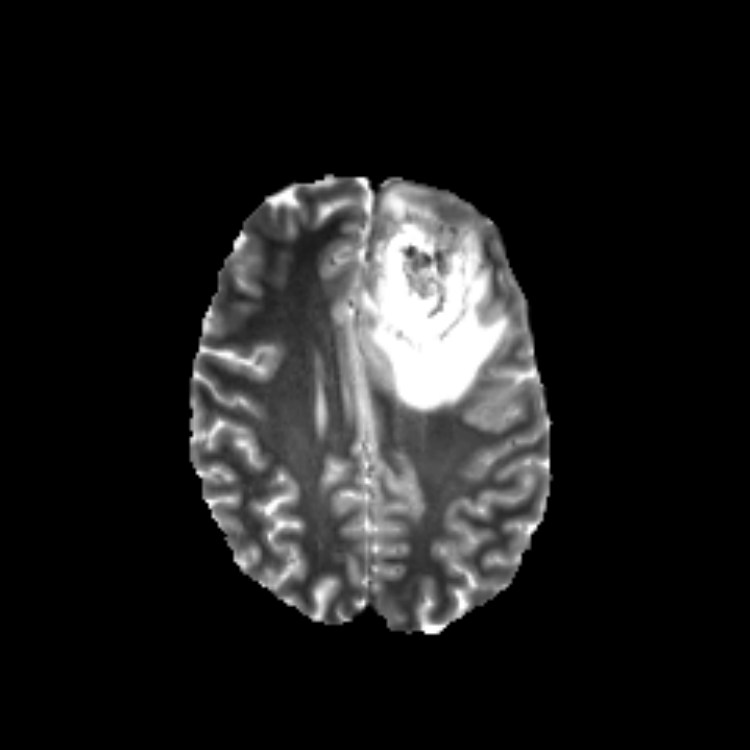}\\
\raisebox{0.09\textwidth}{\rotatebox[origin=c]{90}{\Tone}} &
\synfigure{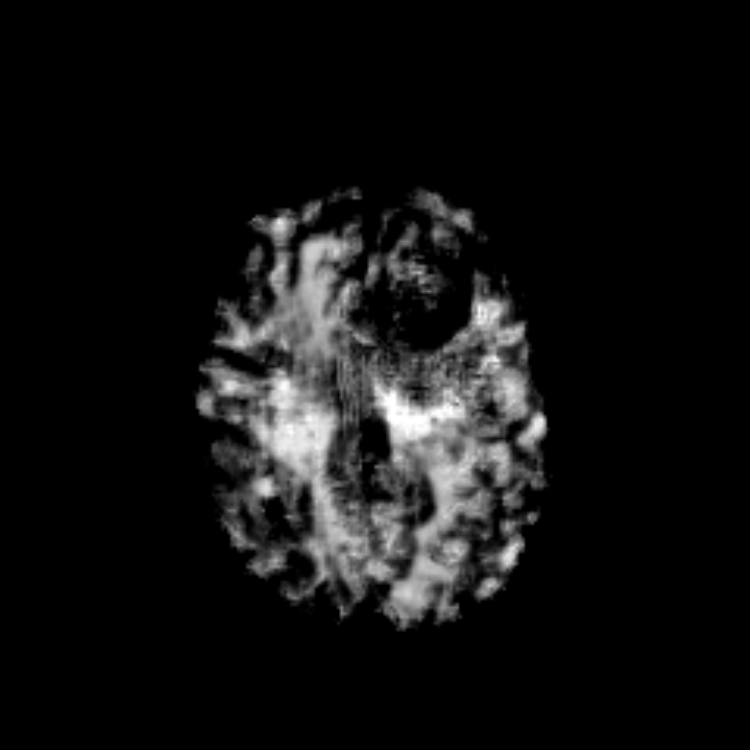} &
\synfiguresyntetic{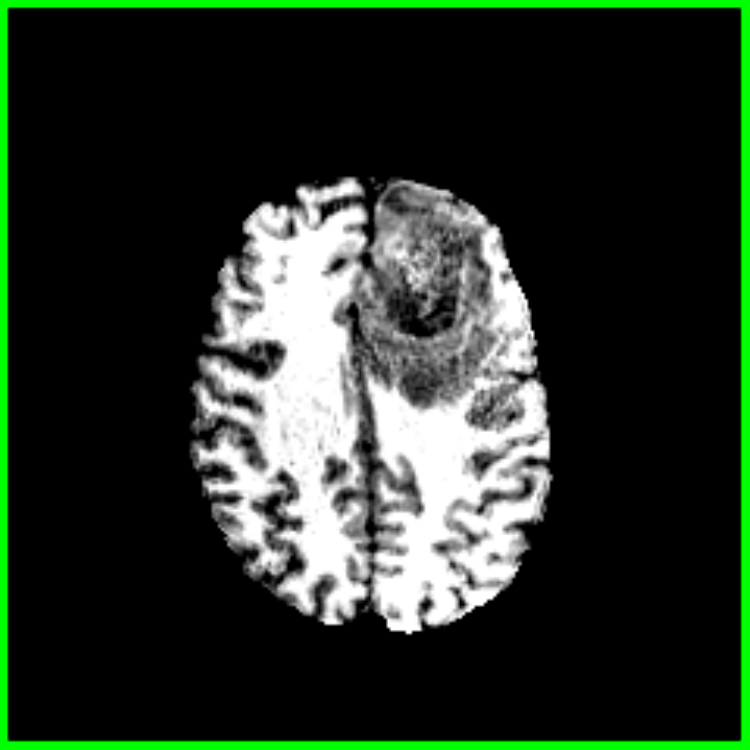} &
\synfiguresyntetic{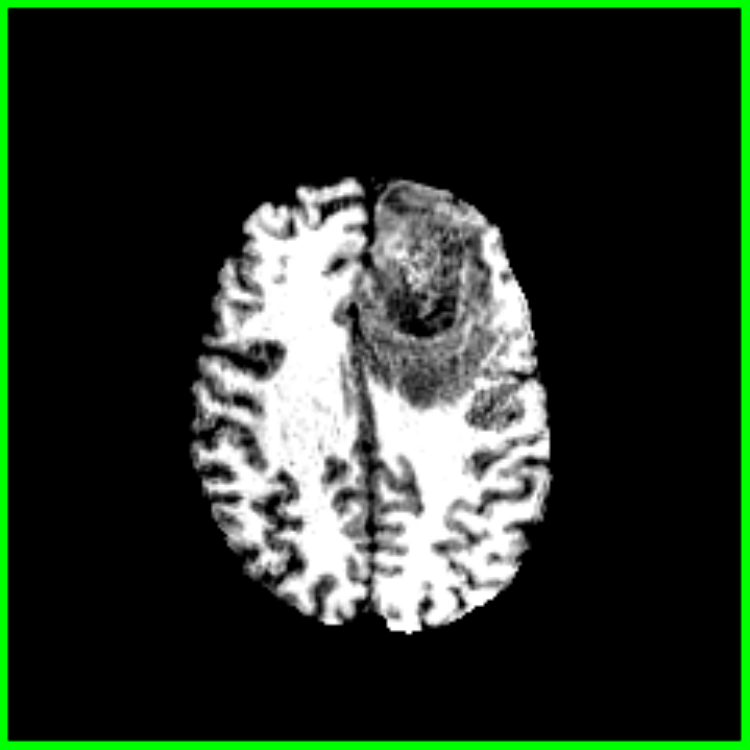} &
\synfiguresyntetic{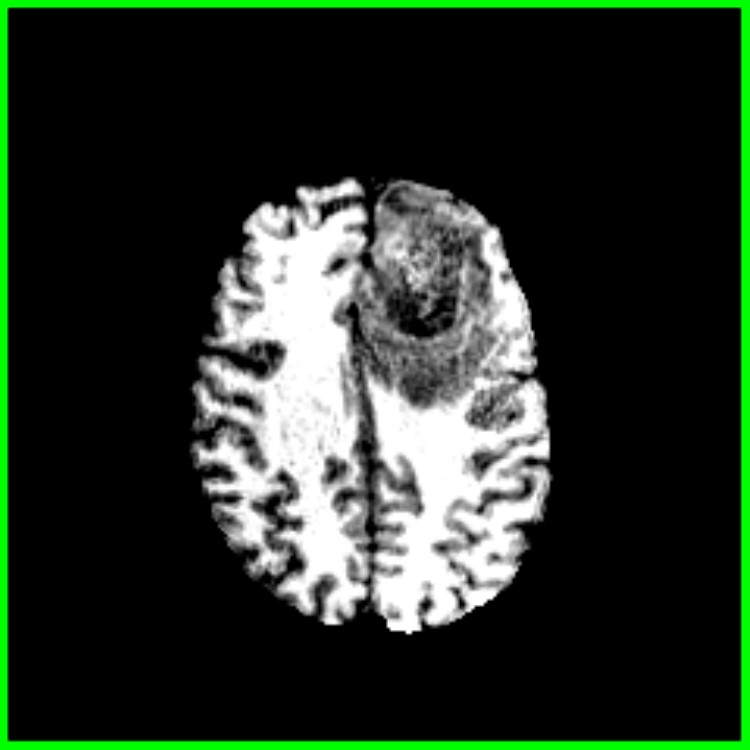} &
\synfigure{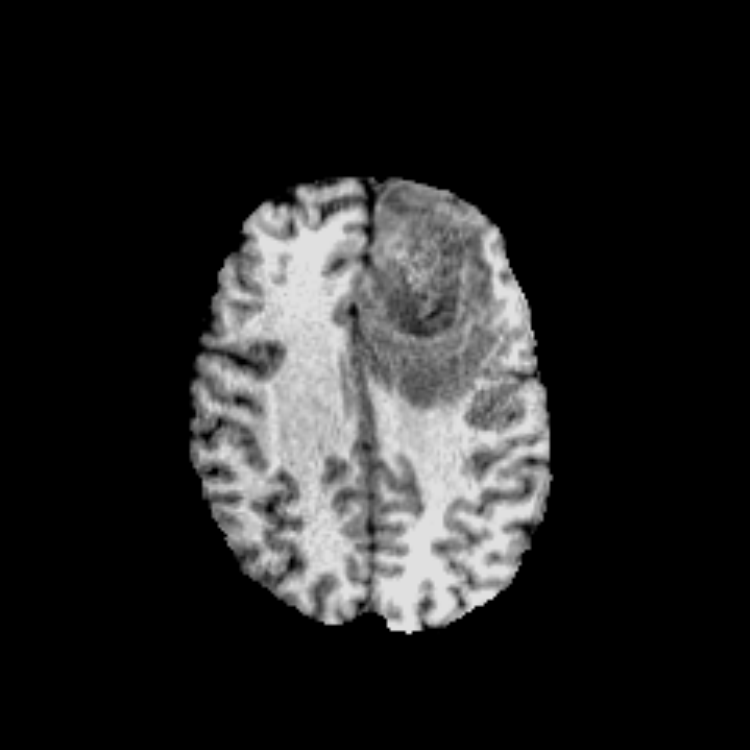}\\
\raisebox{0.09\textwidth}{\rotatebox[origin=c]{90}{\Tonec}} &
\synfiguresyntetic{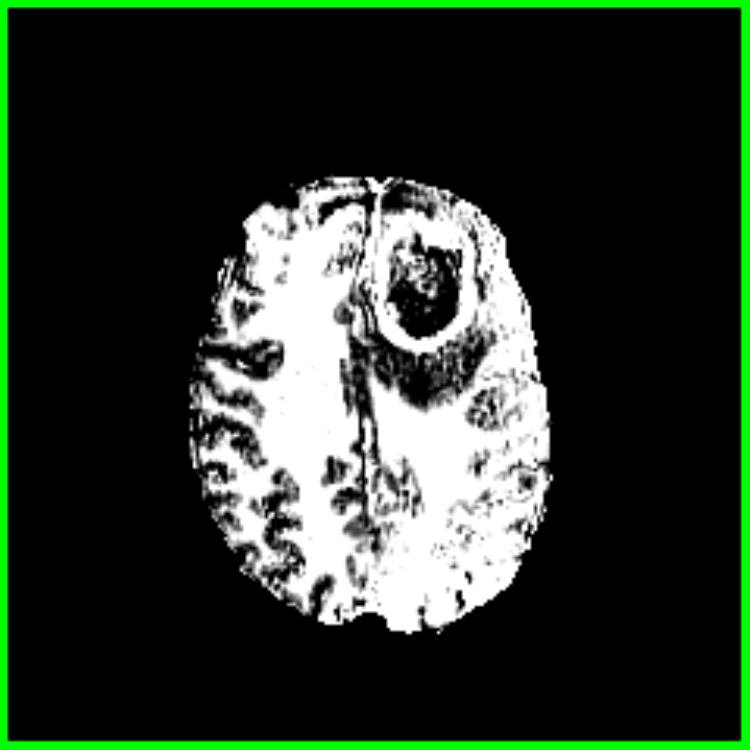} &
\synfiguresyntetic{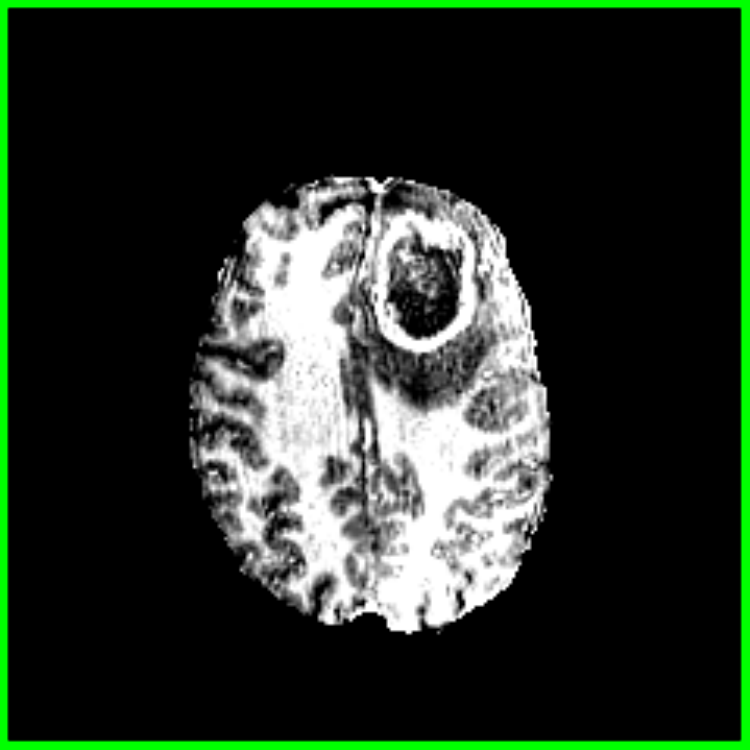} &
\synfiguresyntetic{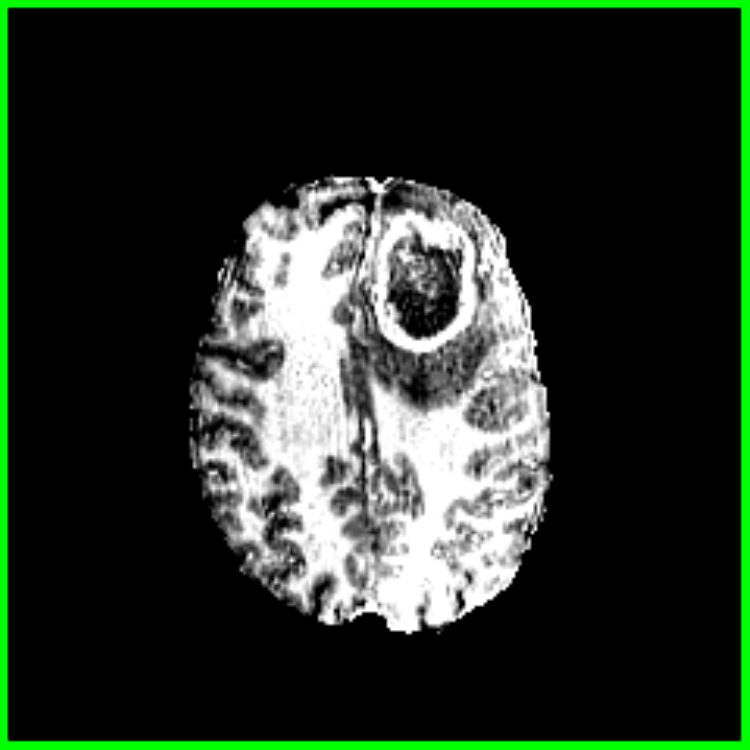} &
\synfiguresyntetic{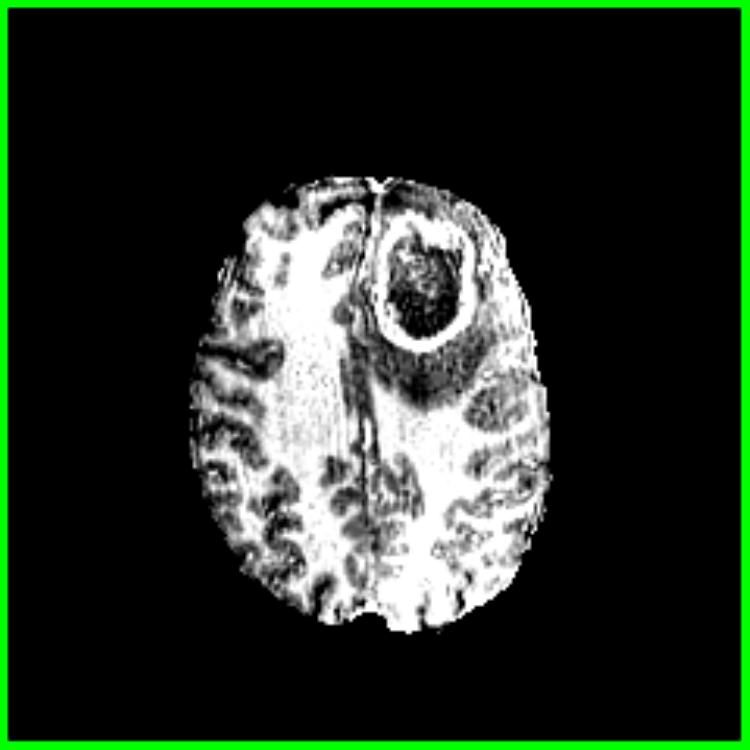} &
\synfigure{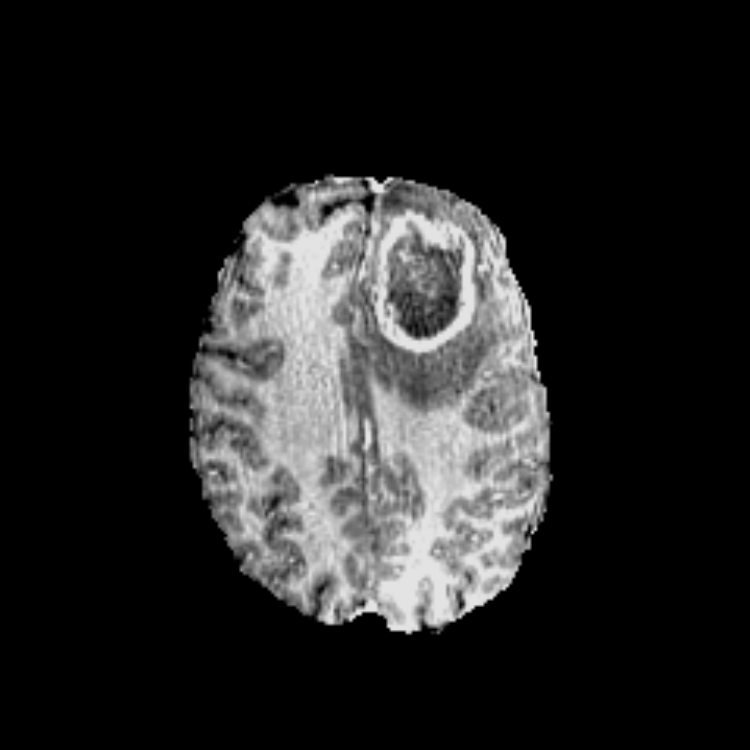}\\
\end{tabular}
\caption{Examples of image reconstruction and synthesis given different combinations of inputs only trained on BRATS in pre-training.
The images with green border are reconstructed with given inputs while those without are synthesized.}
\label{fig:URN_synthesis}
\end{figure}

\begin{figure}[t]
\centering
\begin{tabular}{lccccc}
&
\Tonec &
\Tone, \Tonec &
\Tone, \Tonec, \Ttwo &
All &
GT \\
\raisebox{0.09\textwidth}{\rotatebox[origin=c]{90}{\centering FLAIR}} &
\synfigure{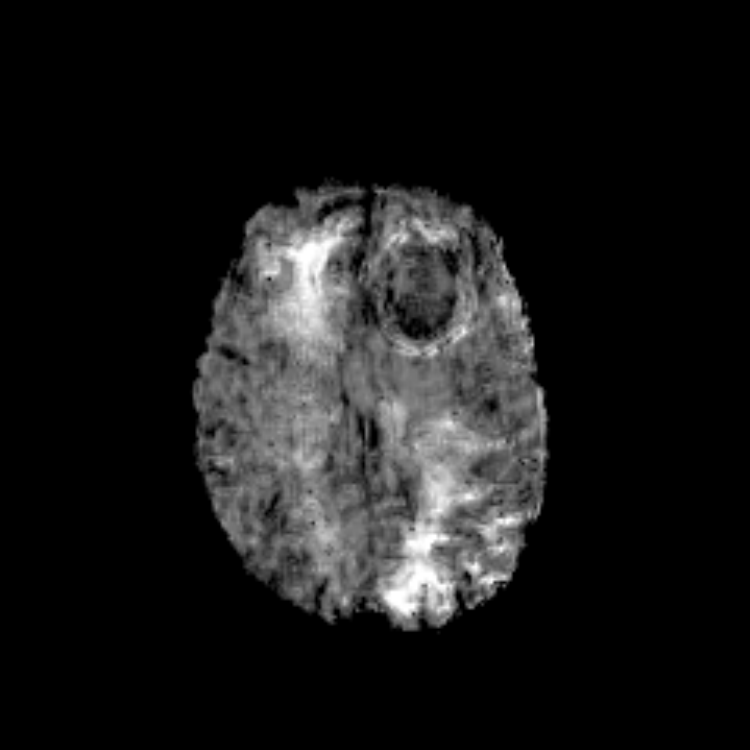} &
\synfigure{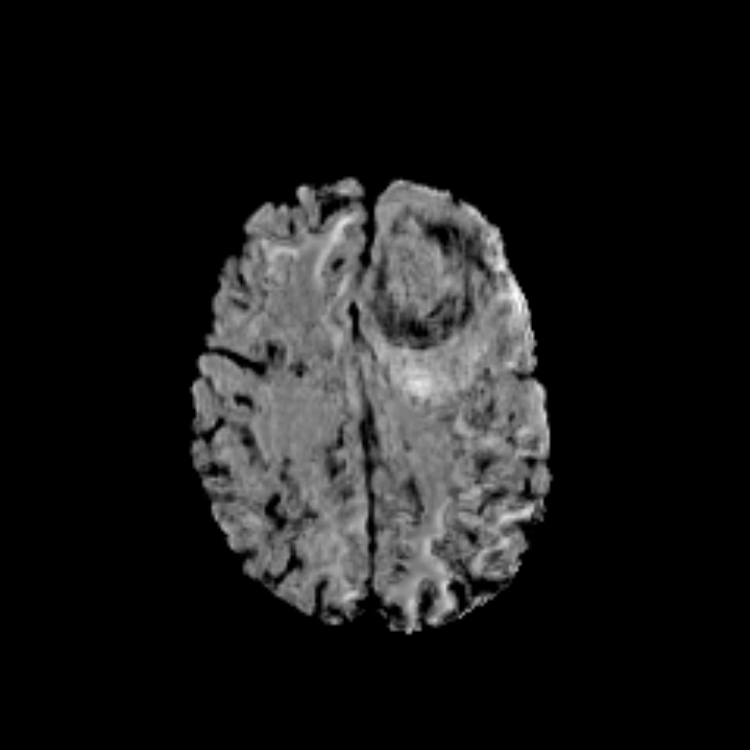} &
\synfigure{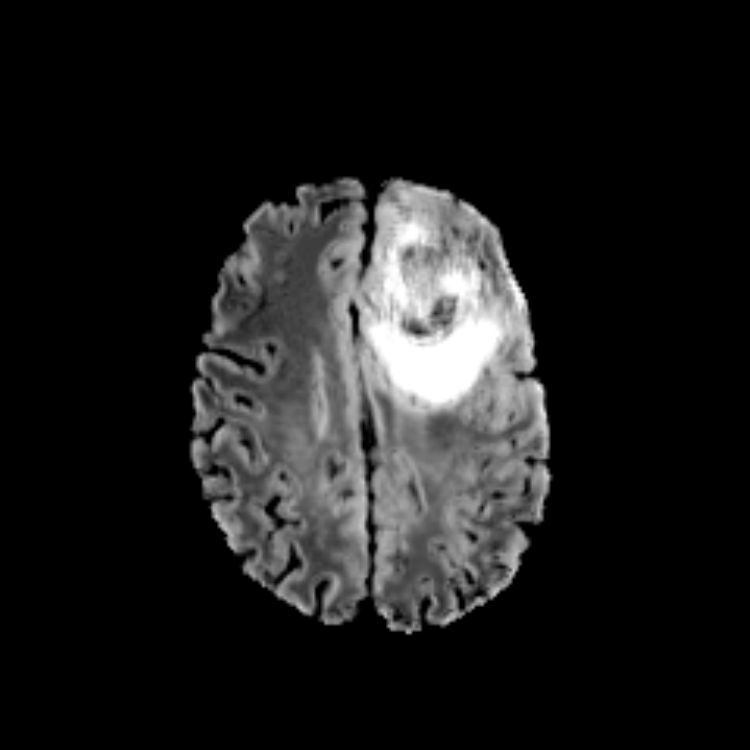} &
\synfiguresyntetic{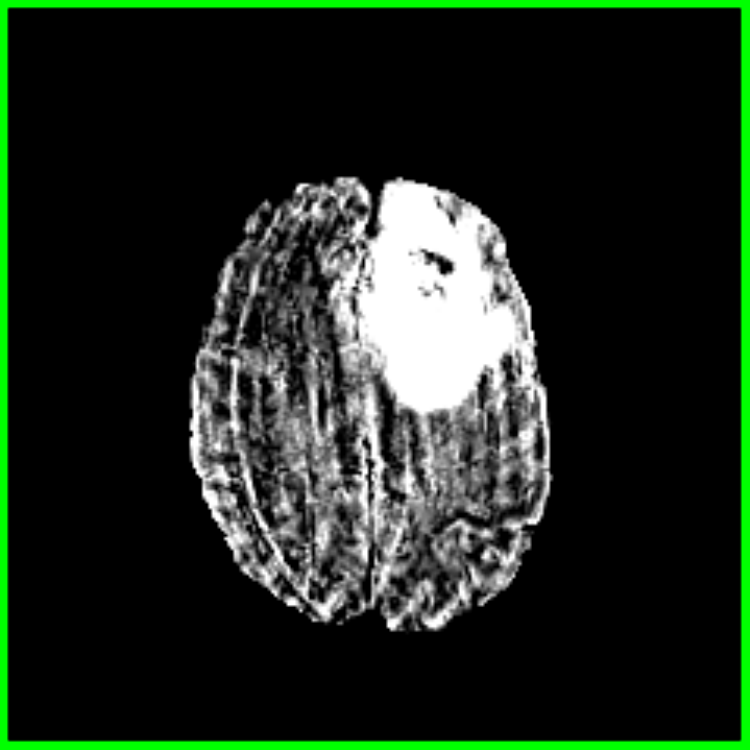} &
\synfigure{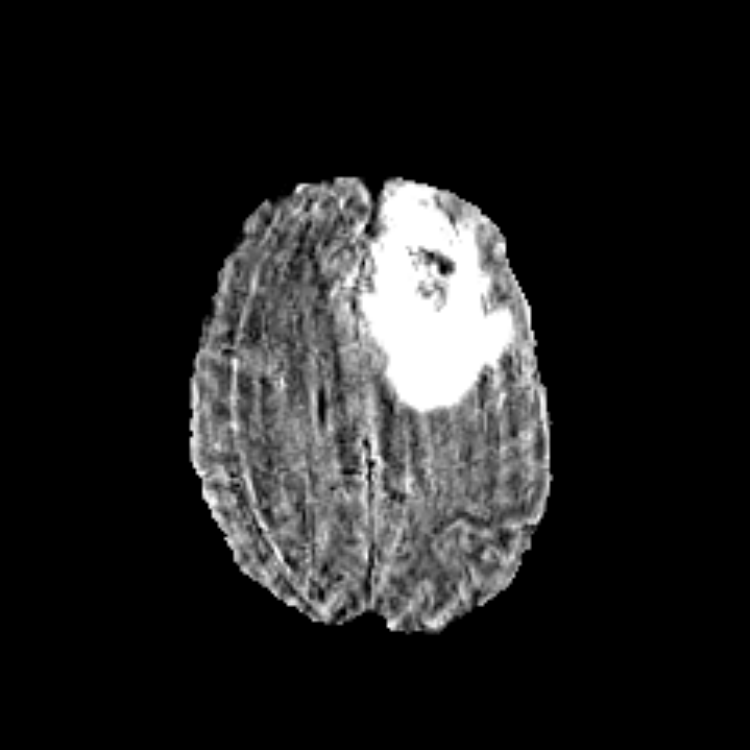}\\
\raisebox{0.09\textwidth}{\rotatebox[origin=c]{90}{\Ttwo}} &
\synfigure{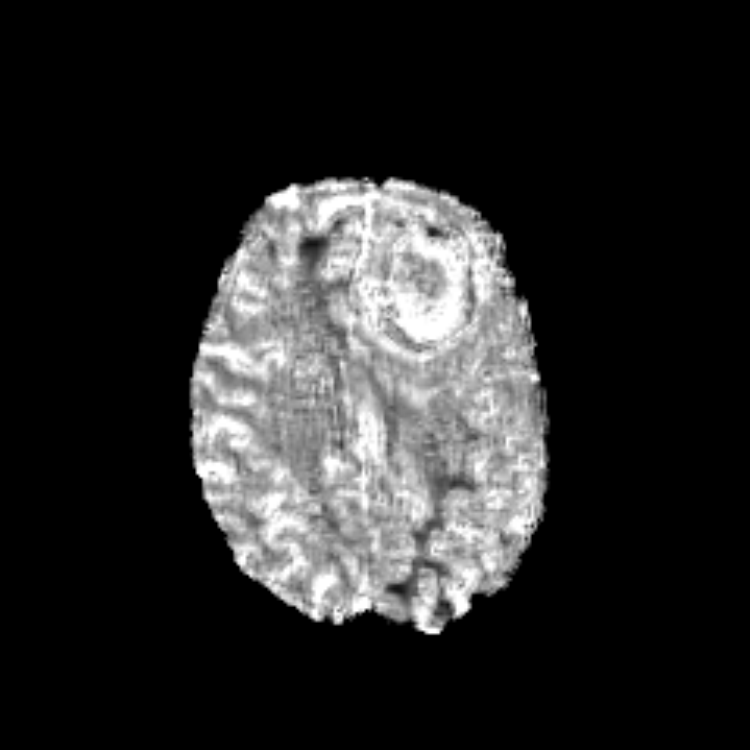} &
\synfigure{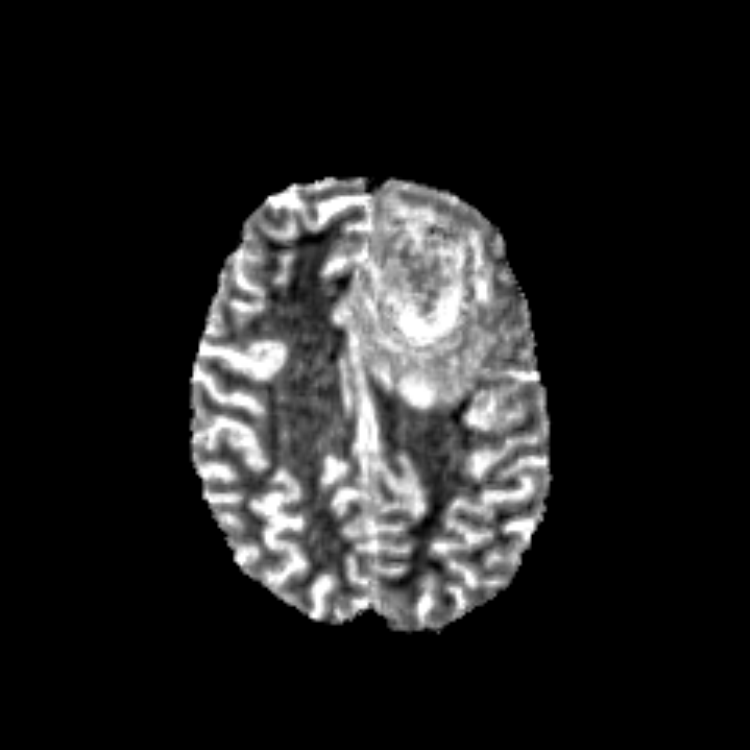} &
\synfiguresyntetic{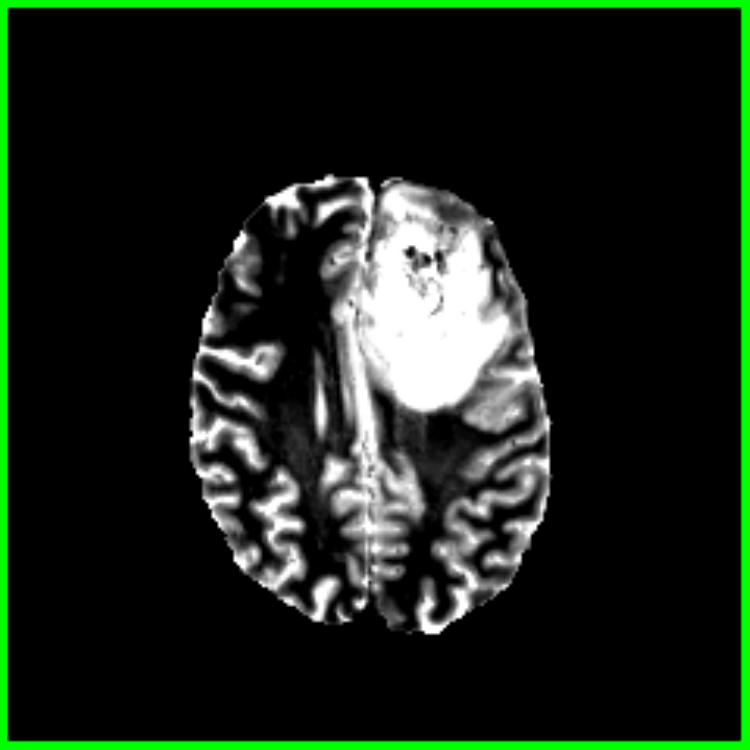} &
\synfiguresyntetic{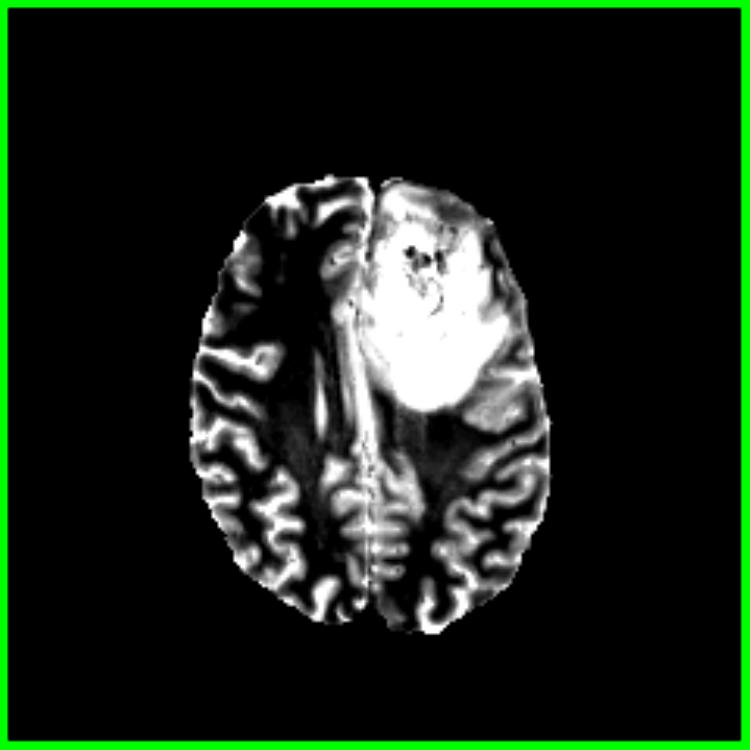} &
\synfigure{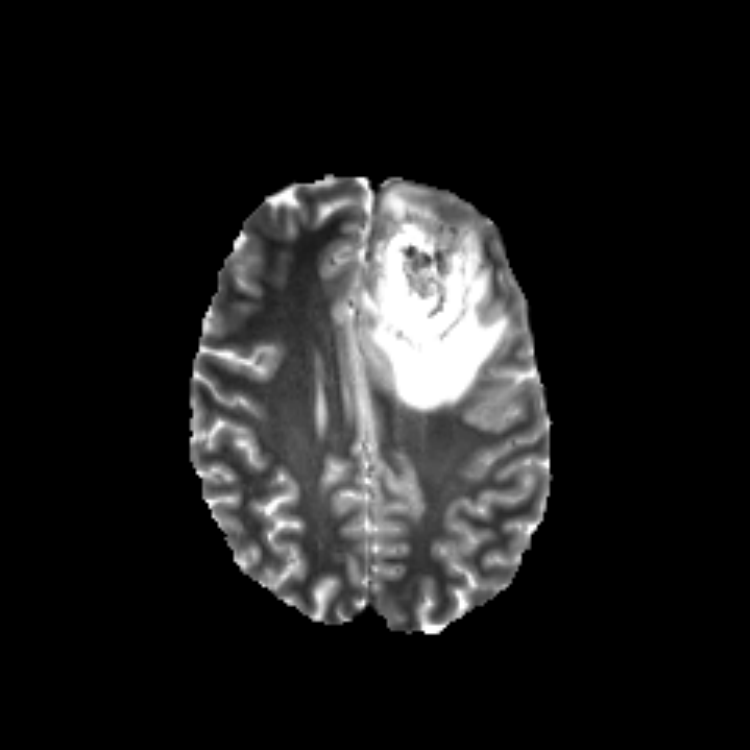}\\
\raisebox{0.09\textwidth}{\rotatebox[origin=c]{90}{$T_1$}} &
\synfigure{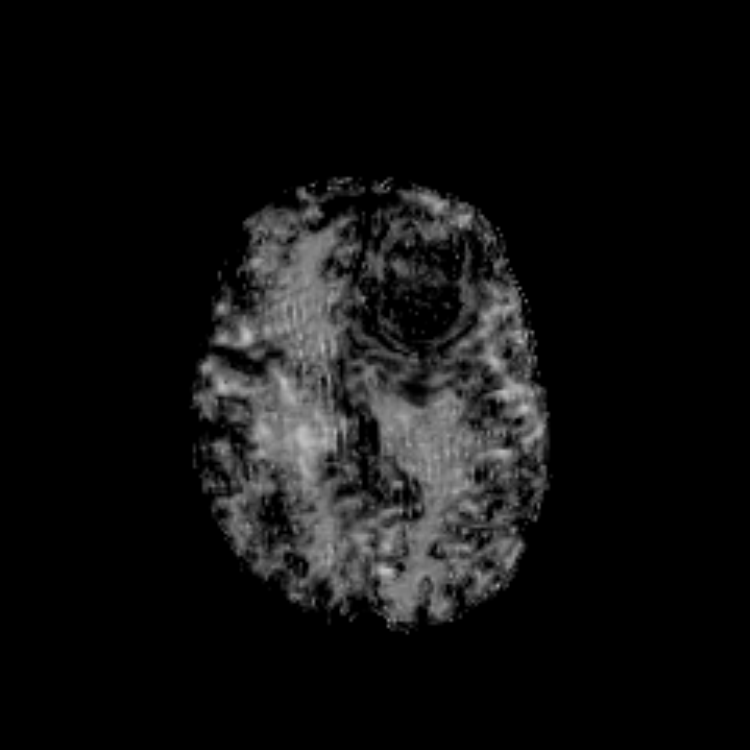} &
\synfiguresyntetic{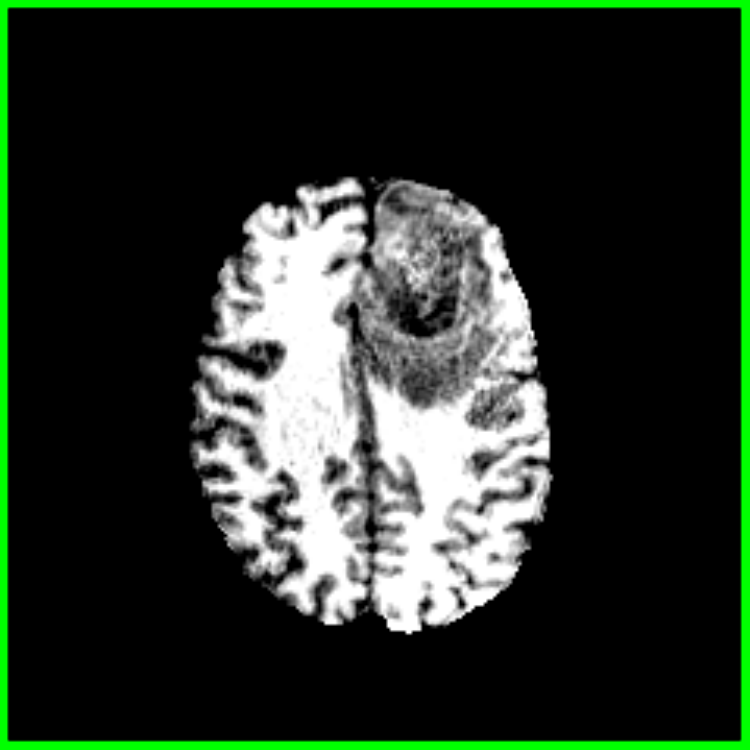} &
\synfiguresyntetic{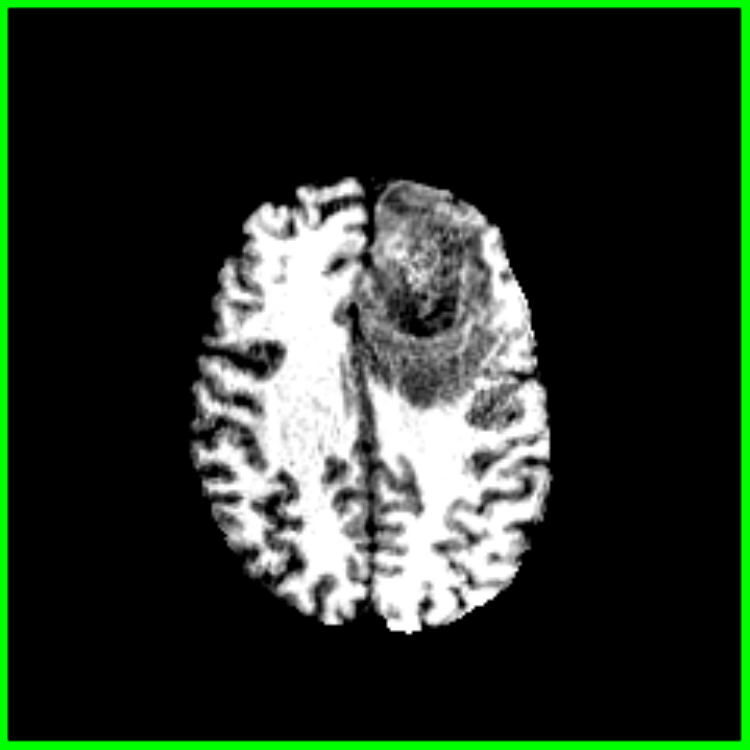} &
\synfiguresyntetic{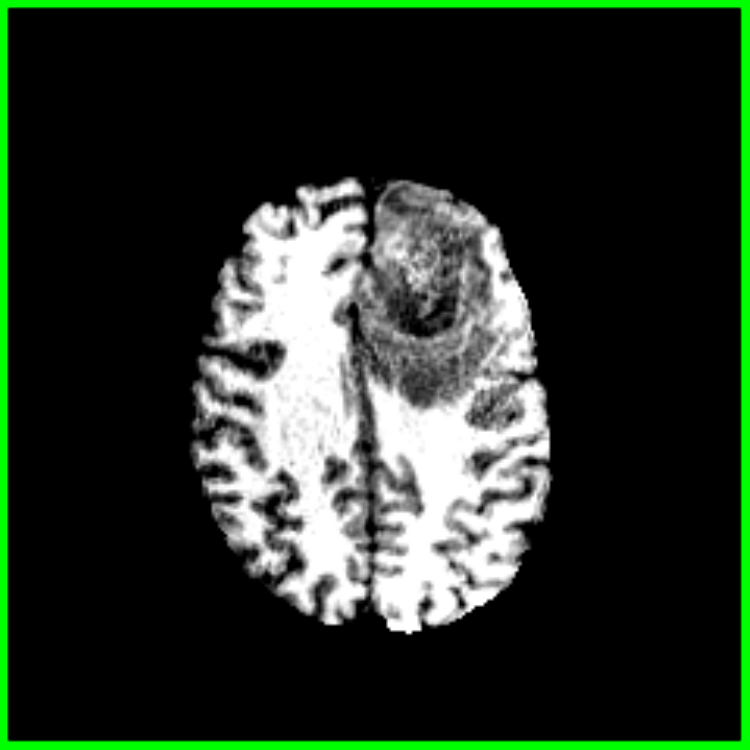} &
\synfigure{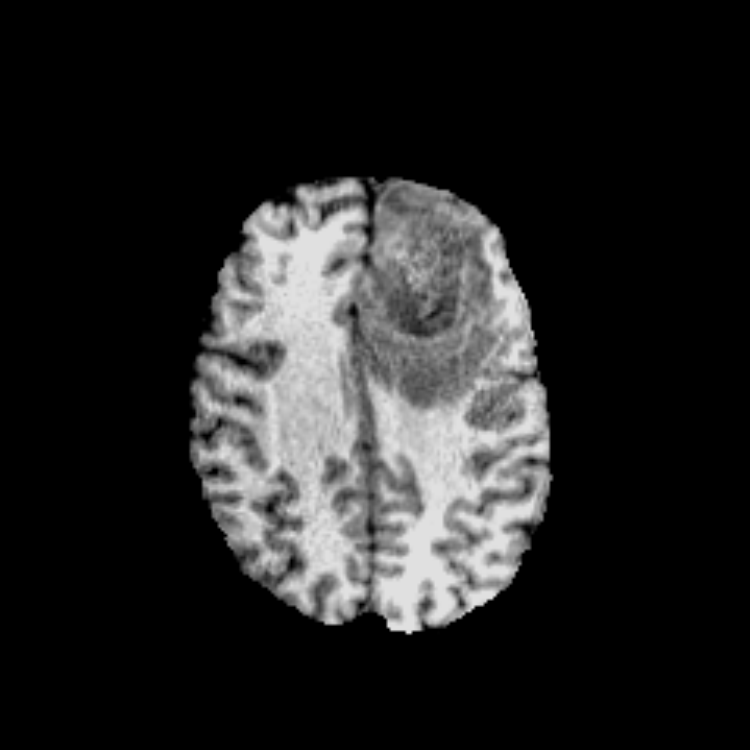}\\
\raisebox{0.09\textwidth}{\rotatebox[origin=c]{90}{\Tonec}} &
\synfiguresyntetic{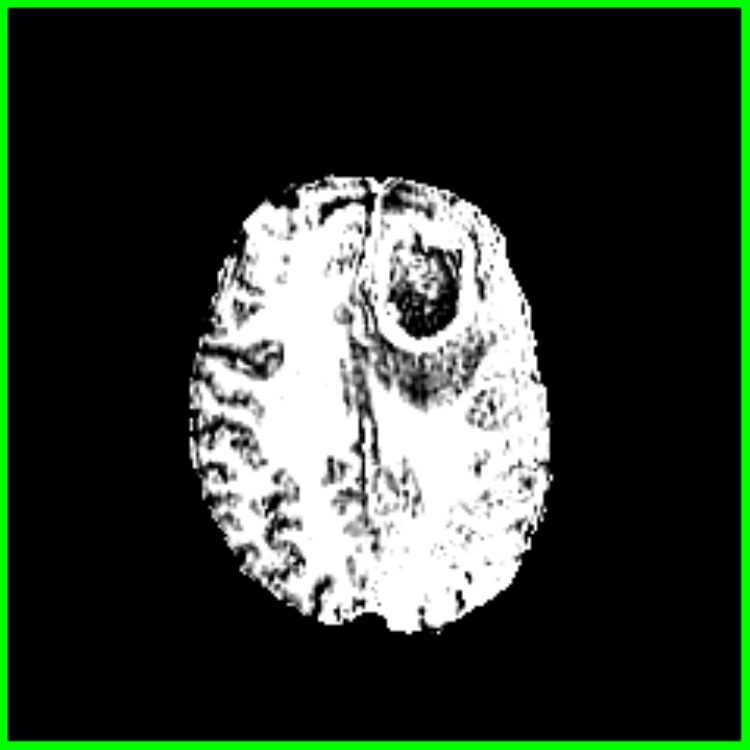} &
\synfiguresyntetic{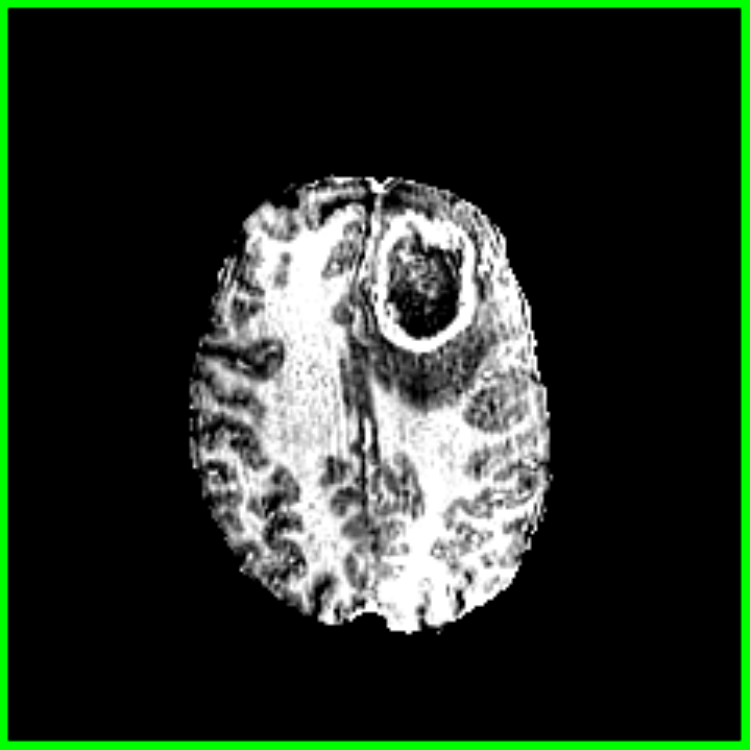} &
\synfiguresyntetic{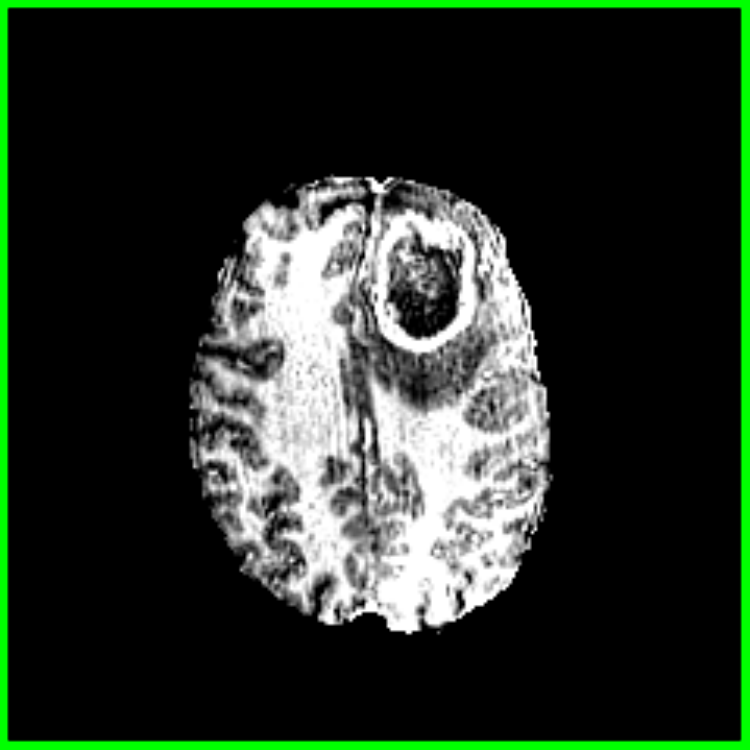} &
\synfiguresyntetic{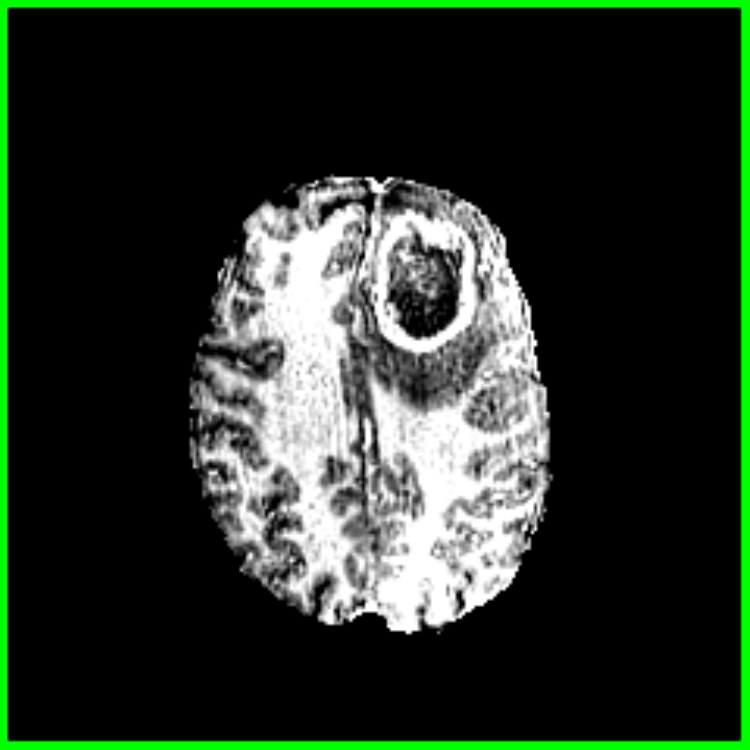} &
\synfigure{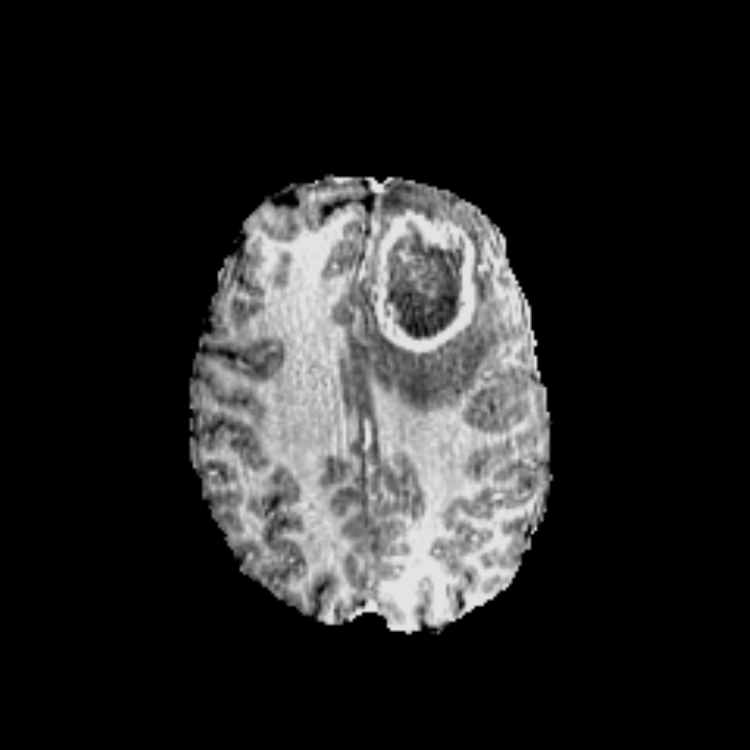}\\
\end{tabular}
\caption{Examples of image reconstruction and synthesis given different combinations of inputs while pre-trained with BRATS and HCP dataset.
The images with green border are reconstructed with given inputs while those without are synthesized.}
\label{fig:URN_wHCP_synthesis}
\end{figure}

\end{document}